\documentclass[journal,twoside,web]{ieeecolor}
\usepackage{generic}
\usepackage{cite}
\usepackage{amssymb,amsfonts}
\usepackage{algorithmic}
\usepackage{graphicx}
\usepackage{algorithm,algorithmic}
\usepackage{hyperref}
\usepackage{textcomp}

\usepackage{amsmath}

\usepackage{epsfig} 
\usepackage{mathptmx} 
\usepackage{times} 
\usepackage{multicol}

\newtheorem{theorem}{Theorem}
\newtheorem{corollary}{Corollary}
\newtheorem{definition}{Definition}
\newtheorem{proposition}{Proposition}
\newtheorem{assumption}{Assumption}
\newtheorem{lemma}{Lemma}

\newcommand{\bR}{\mathbb{R}}
\newcommand{\bP}{\mathbb{P}}
\newcommand{\bE}{\mathbb{E}}
\usepackage{forest}

\usepackage{mathtools}

\newcommand{\calF}{{\mathcal{F}}}
\newcommand{\calN}{{\mathcal{N}}}
\newcommand{\calK}{{\mathcal{K}}}

\newcommand{\calI}{{\mathcal{I}}}

\DeclarePairedDelimiter\norm{\lVert}{\rVert}%

\def\BibTeX{{\rm B\kern-.05em{\sc i\kern-.025em b}\kern-.08em
    T\kern-.1667em\lower.7ex\hbox{E}\kern-.125emX}}
\begin{document}

\title{Exact Recovery for System Identification with More Corrupt Data than Clean Data}
\author{Baturalp Yalcin, Haixiang Zhang, Javad Lavaei, and Murat Arcak
\thanks{This work was supported by grants from AFOSR, ARO, ONR, and NSF.}
\thanks{B. Yalcin and J. Lavaei are with the Department of
Industrial Engineering and Operations Research, University of California, Berkeley, CA, USA, 94720 (e-mail: baturalp\_yalcin@berkeley.edu; lavaei@berkeley.edu). }
\thanks{H. Zhang is with the Department of
Mathematics, University of California, Berkeley, CA, USA, 94720 (e-mail: haixiang\_zhang@berkeley.edu). }
\thanks{M. Arcak is with the Department of Electrical Engineering and Computer Sciences, University of California, Berkeley, CA, USA, 94720 (e-mail: arcak@berkeley.edu).}}

\maketitle

\begin{abstract}

This paper investigates the system identification problem for linear discrete-time systems under adversaries and analyzes two lasso-type estimators. We examine both asymptotic and non-asymptotic properties of these estimators in two separate scenarios, corresponding to deterministic and stochastic models for the attack times. Since the samples collected from the system are correlated, the existing results on lasso are not applicable. We prove that when the system is stable and attacks are injected periodically, the sample complexity for exact recovery of the system dynamics is linear in terms of the dimension of the states. When adversarial attacks occur at each time instance with probability $p$, the required sample complexity for exact recovery scales polynomially in the dimension of the states and the probability $p$. This result implies almost sure convergence to the true system dynamics under the asymptotic regime. As a by-product, our estimators still learn the system correctly even when more than half of the data is compromised. We highlight that the attack vectors are allowed to be correlated with each other in this work, whereas we make some assumptions about the times at which the attacks happen. This paper provides the first mathematical guarantee in the literature on learning from correlated data for dynamical systems in the case when there is less clean data than corrupt data.
\end{abstract}


\section{Introduction}
\label{sec:introduction}

Dynamical systems serve as the fundamental components in reinforcement learning and control systems. The system dynamics may not be known exactly when the system is complex. Therefore, learning the underlying system dynamics, named the system identification problem, and using the data collected from the system are essential in robotics, control theory, time-series, and reinforcement learning applications. The system identification problem with small disturbances using the least square estimator has been ubiquitously studied, and the literature for this problem is overly rich \cite{chen2012identification}. Despite several advances in this field, most results in system identification focus on the asymptotic properties, i.e., properties of the estimators at infinity, of the proposed estimators only. Nonetheless, the non-asymptotic analysis of the system identification problem has gained interest in recent years \cite{hazan2018spectral, mania2019certainty, sarkar2020nonparametric,tsiamis2022statistical}. Although non-asymptotic analysis is harder, it is crucial to understand the required sample complexity for online control problems.

The robust learning of dynamical systems is crucial for safety-critical applications, such as autonomous driving \cite{Alan2022ControlBF}, unmanned aerial vehicles \cite{Wang2017SafeLO}, and robotic arms \cite{KhansariZadeh2014LearningCL}. While recent papers have addressed online non-asymptotic control of linear time-invariant (LTI) systems, their applicability often hinges on the assumption of small noise in measurements, neglecting scenarios involving large magnitudes of noise indicative of adversarial attacks or data corruption \cite{simchowitz2018learning, simchowitz2021naive, zhang2021regret}. These papers utilize recent advances in high-dimensional statistics and learning theory to analyze the properties of the solution even when the data samples are correlated. The work \cite{ziemann2023tutorial} provides a tutorial on proof techniques. Least-square estimators are the main tool in those works, which are susceptible to outliers and large noise in the system. Consequently, we propose two new non-smooth estimators inspired by the lasso problem and robust regression literature \cite{bakoAnalysisNonsmoothOptimization2016}. We study the required sample complexity for the exact recovery of LTI systems using these estimators when there are sporadic large disturbance injections to the system.

The robust regression and learning problems under adversaries are ubiquitously studied in the literature \cite{xuRobustnessRegularizationSupport2009, bakoClassOptimizationBasedRobust2017, bertsimasCharacterizationEquivalenceRobustification2018, pesme2020online}. However, existing methods for analyzing the estimators cannot be directly generalized to control problems due to the correlation between the samples. Therefore, different strategies have been developed recently to tackle this challenge. Firstly, the system is initiated multiple times, and the data point at the end of each run is used to obtain uncorrelated data points, as in \cite{dean2018sample}. However, obtaining multiple trajectories is not viable and cost-efficient for most safety-critical applications. One method with a single trajectory relies on the persistent excitation of the states so that the dynamics can be explored thoroughly. This is achieved by injecting a Gaussian noise input into the system. Small ball techniques are used to analyze the properties of the estimator \cite{mendelson2014learning, simchowitz2018learning, li2021safe}. This technique employs normalized martingale bounds for the estimation error when the excitation is large enough \cite{simchowitz2018learning}.

Unlike the non-asymptotic analysis of correlated data, the least-squares estimator offers a closed-form solution when the system is subjected to small white noise \cite{fattahi2019learning,jedra2020finitetime, wagenmaker2020active}. As long as the noise magnitudes are not large, the least-squares estimator performs relatively well. The estimation error asymptotically converges to zero with the optimal rate of $T^{-1/2}$, where $T$ is the number of samples collected from the system \cite{simchowitz2018learning}. However, it is not robust to adversarial attacks, and the literature on robust learning of dynamical systems is limited. The work by \cite{feng2022learningnsp} defines the null space property (NSP) to analyze a lasso-type estimator for the system. It provides necessary and sufficient conditions for exact recovery when NSP is satisfied, which is NP-hard to check. To circumvent the computational complexity, we build upon \cite{feng2022learningnsp} and study robust estimators from a non-asymptotic point of view under standard assumptions, such as the system being stable and the attacks being sub-Gaussian.

\textbf{Contributions:} We study discrete-time linear time-invariant systems of the form $x_{i+1} = \bar A x_i + \bar B u_i + \bar d_i$, where $\bar A \in \mathbb{R}^{n \times n}$ and $\bar B \in \mathbb{R}^{n \times m}$ are unknown matrices of the model. We aim to learn these matrices from the samples $\{ x_i, u_i \}_{i=0}^{T-1}$ of a single initialization of the system when the disturbance vectors $\bar d_i$ are adversarial. Here, the adversarial noise refers to a vector that is designed to deteriorate the performance of the estimator. Thus, the adversarial vectors $\{\bar d_i \}_{i=0}^{T-1}$ can take arbitrarily large finite values, be dependent over time, and can have any undesirable structures. We say that an adversarial attack occurs whenever $\bar d_i$ is non-zero, and we have no information on the value of $\bar d_i$. If $\bar d_i$ is zero, there is no attack or adversary at time $i$. In our setting, we study systems that are not subject to ordinary minor measurement or modeling errors, and instead the non-zero noise or disturbance stems from an adversarial event.

We study two convex estimators based on the minimization of the $\ell_2$ and $\ell_1$ norms of the estimated disturbance vectors, $\sum_{i=0}^{T-1} \|d_i\|_2$ and $\sum_{i=0}^{T-1} \|d_i\|_1$, with the decision variables $A$, $B$, and $\{ d_i \}_{i=0}^{T-1}$ subject to $x_{i+1} = A x_{i} + B u_i + d_i$, given the samples $\{ x_i, u_i \}_{i=0}^{T-1}$:
\begin{equation*}
    \min_{A \in \mathbb{R}^{n \times n}, B \in \mathbb{R}^{n \times m}}\;\, \sum_{t=0}^{T-1}\| x_{t+1} - Ax_t - Bu_t \|_\circ, \quad \circ \in \{1, 2\}.
\end{equation*}
This is equivalent to an empirical risk minimization problem for which the loss function is the $\ell_1$ and $\ell_2$ norms depending on the choice of $\circ$. We employ a non-smooth objective function to obtain a robust estimator. The arbitrary injection of adversaries may happen infrequently in time. In that case, the attacks occur sparsely in time. Conversely, the vector $\bar d_i$ at each attack time $i$ could be dense, and there is no limitation on how sparse the vector is. The $\ell_2$ norm estimator is the most effective in this case. In contrast, the $\ell_1$ norm estimator is preferable if the vector $\bar d_i$ at each attack time is structured and known to be sparse. We summarize our contributions below.

\textbf{i)} We first consider the case when the adversarial noise injections, i.e., adversarial attacks, happen periodically over time with the period $\Delta$. We show that both of our estimators exactly recover the true system matrices $\bar A$ and $\bar B$ when the system is stable and the number of samples, i.e., $T$, is larger than $n + \Delta$.

\textbf{ii)} We then consider a probabilistic model for the occurrence of attacks, in which there is an arbitrary noise injection at each time instance $i$ with probability $p$, independent of previous time periods. Nevertheless, we allow these noise injections, or attack vectors, to be dependent. We study the required sample complexity of our estimators for exact recovery when the attack vectors are stealthy. 
Suppose that the adversarial noise and the input sequence are sub-Gaussian random vectors and possibly dependent.  Then, the estimators achieve exact recovery with probability at least $1-\delta$ if the time horizon $T$ satisfies the inequality $T \geq \Theta(\max\{T_{\text{sample}}^1,T_{\text{sample}}^2\})$, where $T_{\text{sample}}^1$ and $T_{\text{sample}}^2$ are defined as
\[ n^2 R_1 \log\left(\frac{nR_1}{\delta}\right), \]
and 
\[ nmR_2\log\left(\frac{nR_2}{\delta}\right), \]
with the constants $R_1$ and $R_2$ defined in Theorem \ref{thm: general-l2}.

\textbf{iii)} As a corollary to the previous result, we show that the estimators converge to true system matrices almost surely when the attack vectors are stealthy. Otherwise, if the attack vectors are not stealthy, the system operator could detect the abnormalities and stop the system, which is not a desired outcome for the adversarial agent or attacker. This is the first paper that studies the adversarial attack structure for the system identification problem to obtain sample complexity using non-asymptotic analysis techniques. 

This paper is organized as follows. In Sections 2 and 3, we introduce the notations used in the paper and formulate the problem, respectively. In Section 4, we study the convergence and sample complexity properties of our estimators in the case when the system is autonomous. In Section 5, we generalize the results to non-autonomous systems. In Section 6, we demonstrate the results on a biomedical system that models blood sugar levels with the injection of bolus insulin. This work provides the first bound in the literature on sample complexity for dynamical systems under adversaries, and its techniques can be adopted to study other robust online learning problems.

\section{Notation and Preliminaries}

For a matrix $Z$, $\| Z \|_{F}$ denotes the Frobenius norm of a matrix. For a vector $z$, $\| z \|_1$, $\| z \|_2$, and  $\| z \|_\infty$ denote its $\ell_1$, $\ell_2$, and $\ell_\infty$ norms, respectively. Given two functions $f$ and $g$, the notation $f(x) = \Theta[g(x)]$ means that there exist universal positive constants $c_1$ and $c_2$ such that $c_1 g(x) \leq f(x) \leq c_2 g(x) $. The relation $f(x) \lesssim g(x)$ holds if there exists a universal positive constant $c_3$ such that $f(x) \leq c_3 g(x)$ holds with high probability when $T$ is large. The relation $f(x) \gtrsim g(x)$ holds if $g(x) \lesssim f(x)$.
$| S |$ shows the cardinality of a given set $S$.  For two vectors $v$ and $w$, $\langle v, w \rangle$ is the inner product between those vectors in their respective vector space. Furthermore, we use the notation $v \otimes w = vw^T$ to denote the outer product. $\bP(\cdot)$ and $\bE[\cdot]$ denote the probability of an event and the expectation of a random variable. A Gaussian random variable $X$ with mean $\mu$ and covariance matrix $\Sigma$ is written as $X \sim N(\mu, \Sigma)$.  Since we restrict the disturbance vectors to be sub-Gaussian, we formally define them below.
 \begin{definition}[Sub-Gaussian Random Variable \cite{wainwright_2019}]
       A random variable $X \in \bR$ with mean $\mu = \mathbb{E}[X]$ is sub-Gaussian with parameter $\sigma$ if 
        \[ \mathbb{E}[e^{\lambda(X-\mu)}] \leq e^{\lambda^2\sigma^2 /2}, \quad \forall \lambda \in \mathbb{R}.\]
        Moreover, a random vector $X \in \bR^n$ with mean $\mu = \mathbb{E}[X]$ is sub-Gaussian with parameter $\sigma$ if 
        \[ \mathbb{E}[e^{\lambda \langle \nu,X-\mu \rangle}] \leq e^{\lambda^2 \sigma^2/2}, \quad \forall \lambda \in \mathbb{R}, \nu \in \bR^n, \|\nu\|_2 = 1.\]
    \end{definition}
    Informally, a sub-Gaussian random variable with parameter $\sigma$ has the property that its tails are less dense than those of a Gaussian random variable with variance $\sigma^2$. We will utilize concentration bounds for sub-Gaussian random variables to verify that the optimality conditions for our proposed estimators are satisfied with high probability. The main concentration inequality for sub-Gaussian random variables is Hoeffding's bound.
    \begin{lemma}(Hoeffding's Bound \cite{wainwright_2019}) Suppose that the variable $X$ has mean $\mu$ and sub-Gaussian parameter $\sigma$. Then, for all $t > 0$, we have
    \[ \mathbb{P}\left( |X - \mu| > t \right) \leq 2\exp\left( -\frac{t^2}{2 \sigma^2}  \right).\]
    \end{lemma}
    
    We use the union bound over the set of coordinates and other sets with finite cardinality. Let $S$ be a set with finite cardinality, $|S| < \infty$, and $E_i$ be the event related to element $i$ in the set $S$. Then, we can write the union bound as
    \[ \mathbb{P}\left( \cup_{i \in S} E_i \right) \le \sum_{i \in S} \mathbb{P}(E_i).   \]
    Since we use non-smooth objective functions with $\ell_1$ and $\ell_2$ norms,  we introduce the subdifferentials of the $\ell_1$ and $\ell_2$ norms.
 \begin{definition}[Subdifferential of $\ell_2$ Norm]
     Given a vector $z \in \bR^n$, the subdifferential of $\|z\|_2$ is denoted as $\partial \|z\|_2$ and is given as 
     \begin{align*}
         \partial \|z\|_2 = \begin{cases}
         \frac{z}{\|z\|_2}, & \quad \text{if } z \not = 0, \\ 
         \mathbb{B}_2(1), & \quad \text{ otherwise.}
     \end{cases}
    \end{align*}
     where $\mathbb{B}_2(1) = \{ x \in \bR^n : \|x\|_2 \le 1\}$ is the $\ell_2$ norm unit ball.
 \end{definition}
 
  \begin{definition}[Subdifferential of $\ell_1$ Norm]
        Given a vector $z \in \bR^n$ with entries $z_i, i= 1, \dots, n$, the subdifferential of the $\|z\|_1$ is denoted as $\partial \|z\|_1$ and is given as 
     \begin{align*}
         \partial \|z\|_1^i = \begin{cases}
         1, & \quad \text{if } z_i > 0, \\ 
         -1, & \quad \text{if } z_i < 0, \\ 
         [-1,1], & \quad \text{ otherwise}, 
     \end{cases}
    \end{align*}
     where $\partial \|z\|_1^i$ is the $i$-th coordinate of the subdifferential of $\|z\|_1$.
 \end{definition}

Note that while the subdifferential of the $\ell_1$ norm is coordinate-wise separable, the subdifferential of the $\ell_2$ norm is not coordinate-wise separable. Whenever the vector $z$ is equal to $0$, the subdifferential of the $\ell_2$ norm is the $\ell_2$ norm unit ball, whereas the subdifferential of the $\ell_1$ norm is the $\ell_\infty$ norm unit ball, which is 
 \[ \mathbb{B}_\infty(1) = \{ x \in \bR^n : \|x \|_\infty \le 1 \}. \]
 We also define the unit ball $\mathbb{S}_2(1)$ as
        \[ \mathbb{S}_2(1) = \{ x \in \bR^{n} : \|x\|_2 = 1 \}\]
 that is the set of all the points on the sphere with radius $1$.
 
The asymptotic analysis of the system identification problem concerns the convergence rate to the true parameter at an infinite-time horizon. However, historically, asymptotic analysis has not provided the required sample complexity to obtain a solution within a given error tolerance. In contrast, non-asymptotic analysis deals with the finite-time behavior of the estimators using learning theory and high-dimensional statistics. It provides the required sample complexity to bound the estimation error within the specified tolerance with high probability. Consequently, non-asymptotic analysis is more challenging than asymptotic analysis. Our goal is to provide the minimum required number of samples to recover the true parameters of the system with a high probability using techniques designed for non-asymptotic analysis.

 \section{Problem Formulation}

We consider a linear time-invariant dynamical system over the time horizon $[0, T]$, $x_{i+1} = \bar A x_i + \bar B u_i + \bar d_i, i= 0, 1, \dots, T-1$,
where $\bar A \in \mathbb{R}^{n \times n}$ and $\bar B \in \mathbb{R}^{n \times m}$ are unknown system matrices, and $\bar d_i \in \mathbb{R}^n$ are unknown system disturbances. Given the set of state measurements $\{ x_i\}_{i = 0}^T$ and the set of inputs $\{ u_i \}_{i = 0}^{T-1}$, the goal is to estimate the unknown system matrices $\bar A$ and $\bar B$. In this paper, the disturbance vectors $\{ \bar d_i \}_{i = 0}^{T-1}$ can be engineered to be large if there is an outside attack on the system from an agent or there is a sensor/actuation fault that leads to major corruption in the system dynamics. Throughout the paper, the disturbance vectors $\{ \bar d_i \}_{i = 0}^{T-1}$ are also called (adversarial) attack vectors. Moreover, the agent who engineers the disturbance vectors is called an attacker. As opposed the majority of the literature, we assume that the disturbance vectors $\{ \bar d_i \}_{i = 0}^{T-1}$ can be dependent on the disturbance vectors from the previous time instances and there is no specific distribution assumption for these vectors except the sub-Gaussian assumption. We represent the time indices of the attacks or large disturbance vectors with the set $\calK$, that is $\calK = \{ i : \bar d_i \not = 0, i \in 0,1, \dots, T-1 \}$. These time instances are called the attack times and $\calK$ is the set of attack times. Similarly, the set of time instances without attack or corrupted data is shown with $\calK^c = \{ i : \bar d_i  = 0, i \in 0,1, \dots, T-1 \}$. These time instances are called the no-attack times, and $\mathcal{K}$ is the set of no-attack times. The data corresponding to attack times are corrupted, whereas the data corresponding to no-attack times are uncorrupted.

We establish the exact recovery of the proposed estimators when there are large disturbances in the system. In such cases, the least-squares method cannot achieve exact recovery, a fact that can be easily verified from its closed-form solution. Define the matrices $X:=[x_0, \ldots, x_{T-1}]$ and $\bar D:=[ \bar d_0, \ldots, \bar d_{T-1}]$. The solution for the least-squares problem is $\hat{A} = (\bar A X + \bar D)^TX(X^TX)^{-1}$ in the absence of the input sequence $\{u_i\}_{i=0}^{T-1}$. Thus, the estimation error is $\|\bar D^T X(X^TX)^{-1}\|$, which is non-zero and arbitrarily large in the presence of arbitrarily large disturbance vectors. A similar calculation can be made in the presence of an input sequence. Consequently, the least-squares estimator cannot achieve a zero estimation error, leading to a plateau in the estimation error of the least-squares estimator in our numerical experiments in Section 6. We define the matrix $D :=[d_0, \ldots, d_{T-1}]$ with its columns being estimated disturbances, as well as the norms of matrices $ \|D\|_{1, 1}:= \sum_i \|d_i\|_1$, and $\|D\|_{2, 1}:= \sum_i \|d_i\|_2$. To exactly recover the system matrices $\bar A$ and $\bar B$, we analyze the following convex optimization problems with non-smooth objective functions:
\begin{align}
  \min_{ \substack{A \in \mathbb{R}^{n \times n}, B \in \mathbb{R}^{n \times m}, \\ D\in \mathbb{R}^{n \times T}} }\;\, & \|D\|_{2, 1}  \label{eq:lasso-2}   \tag{CO-L2}                    \\
  s.t. \quad                   & x_{i+1} = A x_i + B u_i + d_i, \quad {i=0, \ldots, T-1}, \notag
\end{align}
and
\begin{align}
  \min_{ \substack{A \in \mathbb{R}^{n \times n}, B \in \mathbb{R}^{n \times m}, \\ D\in \mathbb{R}^{n \times T} } }\;\, & \|D\|_{1, 1}  \tag{CO-L1} \label{eq:lasso-1}                      \\
  s.t. \quad                   & x_{i+1} = A x_i + B u_i + d_i, \quad {i=0, \ldots, T-1}, \notag
\end{align}
where the states $\{x_i\}_{i=0}^T$ are generated according to $x_{i+1} = \bar A x_i + \bar B u_i + \bar d_i, \quad {i=0, \ldots, T-1}$.
The difference between problems \eqref{eq:lasso-2} and \eqref{eq:lasso-1} is their objective functions. Note that these two problems are equivalent when we have a first-order system with $x_i \in \bR, i \in 0,\dots, T-1$. In problem \eqref{eq:lasso-2}, the sum of the $\ell_2$ norm columns is analogous to the $\ell_1$ norm minimization in the lasso problem. In other words, the $\ell_1$ norm is applied at the group level to $\{d_i\}_{i =0}^{T-1}$ because the occurrence of large injections of disturbances is rare and not frequent. We highlight that the vectors $ \{\bar d_i\}_{i =0}^{T-1}$ are not necessarily sparse. On the other hand, the $\ell_1$ norm is applied both at the group level and the in-group levels to  $\{d_i\}_{i =0}^{T-1}$ for problem \eqref{eq:lasso-1}. For those applications that the disturbance vectors can be assumed to be sparse, \eqref{eq:lasso-1} is more suitable than \eqref{eq:lasso-2}. Furthermore, the states $x_i$ are correlated to each other due to the system dynamics, which makes the non-asymptotic analysis of the problem more challenging than the robust regression literature for which the samples are assumed to be independently generated. One can write the optimization problems \eqref{eq:lasso-2} and \eqref{eq:lasso-1} as follows using the $\ell_2$ and $\ell_1$ norms, respectively:
\begin{equation*}
    \min_{A \in \mathbb{R}^{n \times n}, B \in \mathbb{R}^{n \times m}}\;\, \sum_{t=0}^{T-1}\| x_{t+1} - Ax_t - Bu_t \|_\circ.
\end{equation*}
This is equivalent to an empirical risk minimization problem for which the loss function is the $\ell_1$ or $\ell_2$ norm, depending on the choice of $\circ$. Although these types of sum-of-norm minimization non-smooth loss functions are utilized in other applications, this paper marks the first non-asymptotic analysis of these loss functions in the context of control and system identification with serially correlated data.

We remark that classical statistical theory on empirical risk minimization is not applicable to the problem under study in this paper due to the correlated data at each time instance. By representing the data points $X_t$ as tuples $(x_{t+1}, x_t, \mu_t)$, it is impossible to claim that $X_t$ and $X_{t+1}$ are independent, which is a key assumption in the empirical risk minimization literature. As the first step of our proof technique, the Karush-Kuhn-Tucker (KKT) conditions will be used to analyze the properties of these estimators. Since \eqref{eq:lasso-2} and \eqref{eq:lasso-1} are convex optimization problems with linear equalities, the KKT conditions are necessary and sufficient to guarantee optimality, as stated below.

\begin{theorem}
    Consider the convex optimization problems \eqref{eq:lasso-2} and \eqref{eq:lasso-1} and let $\circ \in \{1,2 \}$. Given a pair of matrices $(\hat A, \hat B)$, if the following conditions hold simultaneously
    \begin{multline}
        0 \in \sum_{i \not \in \mathcal{K}} x_i \otimes \partial \|(\bar A - \hat A)x_i + (\bar B - \hat B)u_i \|_\circ \\ + \sum_{i \in \mathcal{K}} x_i \otimes \partial \|(\bar A - \hat A)x_i + (\bar B - \hat B)u_i + \bar d_i \|_\circ,
    \end{multline}
    \begin{multline}
        0 \in \sum_{i \not \in \mathcal{K}} u_i \otimes \partial \| (\bar A - \hat A)x_i + (\bar B - \hat B)u_i \|_\circ \\ + \sum_{i \in \mathcal{K}} u_i \otimes \partial \|(\bar A - \hat A)x_i  + (\bar B - \hat B)u_i + \bar d_i \|_\circ,
    \end{multline}
    then $(\hat A, \hat B)$ is a solution to \eqref{eq:lasso-1} when $\circ = 1$ and a solution to \eqref{eq:lasso-2} when $\circ = 2$.
    \label{thm: kkt-input}
\end{theorem}
The proof for the KKT conditions when $\circ = 2$ is provided in \cite{han2021}, and the proof for the case $\circ = 1$ can be done similarly. We will utilize the conditions above to study in what scenarios the exact recovery is achievable. As a simple corollary to Theorem \ref{thm: kkt-input}, we can state that $(\bar A, \bar B)$ is a solution to our estimator(s) if the following conditions hold:
    \begin{align*}
            0 \in \sum_{i \not \in \mathcal{K}} x_i \otimes \partial \| 0 \|_\circ+ \sum_{i \in \mathcal{K}} x_i \otimes \partial \|\bar d_i \|_\circ, \\
            0 \in \sum_{i \not \in \mathcal{K}} u_i \otimes \partial \| 0 \|_\circ+ \sum_{i \in \mathcal{K}} u_i \otimes \partial \|\bar d_i \|_\circ .
        \end{align*}

\section{Autonomous Systems}

In this section, we consider autonomous systems, meaning that $u_0 = \cdots = u_{T-1} = 0$. Therefore, the system dynamics could be written as $x_{i+1} = \bar A x_i + \bar d_i$ for $i=0, \ldots, T-1$. Throughout this section, we assume that the system is stable and that it is initialized at the origin.
\begin{assumption}\label{asp:stable}
    Given an autonomous system $x_{i+1} = \bar A x_i + \bar d_i$ for $i=0, \ldots, T-1$ with dimension $n$, assume that $x_0 = 0$ and all eigenvalues of $\bar A$ are inside the unit circle.
\end{assumption}

The stability assumption is standard in system identification problems to avoid an unbounded growth of the states during the learning process. Without loss of generality, we initialize the trajectories at the origin since an initialization at other points affects the results only with a constant factor. We study noiseless systems under an adversary to obtain exact recovery results, meaning that if there is no attack at time $i$, $i \in \calK^c$, then $\bar d_i = 0$. 

In the noisy case, one can consider the following setup. If there is no attack at time $i$, $i \in \calK^c$, then $\bar d_i$ is likely non-zero with a small variance and its value is independent of those for other time periods. If there is an attack at time $i \in \calK$, then $\bar d_i$ is a combination of two terms: a small noise vector that is Gaussian and independent of past time periods, and a large noise vector that could have an arbitrary distribution and possibly be dependent on past time instances. The noisy case, where the system is subjected to small independent and identically distributed Gaussian errors due to measurements and modeling errors, in addition to the adversarial vectors, can be easily addressed using our framework. The perturbation analysis allows us to bound how far the recovered solution is from the true solution in terms of the values of small noise vectors.

Therefore, we only study the noiseless case as described above. Thus, we are interested in recovering the system matrix $\bar{A}$ using the following convex optimization problems for autonomous systems:

\begin{equation}
    \begin{aligned}
        \min_{\substack{A \in \mathbb{R}^{n \times n}, \\ D\in \mathbb{R}^{n \times T}} }\;\, & \sum_{i=0}^{T-1} \norm{d_i}_2                            \\
    s.t. \quad                & x_{i+1} = A x_i + d_i,
    \end{aligned} \label{eq:hard-lasso2} \tag{CO-L2-Aut}
\end{equation}
and
\begin{equation}
    \begin{aligned}
        \min_{\substack{A \in \mathbb{R}^{n \times n}, \\ D\in \mathbb{R}^{n \times T}} }\;\, & \sum_{i=0}^{T-1} \norm{d_i}_1                            \\
    s.t. \quad                & x_{i+1} = A x_i + d_i.
    \end{aligned} \label{eq:hard-lasso1} \tag{CO-L1-Aut}
\end{equation}
The optimality conditions for problem \eqref{eq:hard-lasso2} with $\circ= 2$ and problem \eqref{eq:hard-lasso1} with $\circ= 1$ can be written as follows using Theorem \ref{thm: kkt-input}:
\begin{align}
             0 \in \sum_{i \not \in \mathcal{K}} x_i  \otimes \partial \|(\bar A - A)x_i\|_\circ + \sum_{i \in \mathcal{K}} x_i \otimes \partial \|((\bar A - A)x_i + \bar d_i)\|_\circ. \label{eq: kkt2}
\end{align}
As a remark, although the set of attack times $\mathcal{K}$ appears in the optimality conditions, this set is not known a priori to the system operator. The set is only used during the analysis of the proposed estimators to derive sufficient conditions for exact recovery.

We first consider first-order systems where $x_i, \bar d_i \in \mathbb{R}, i=0,1,\dots,T-1$ and $\bar A \in \mathbb{R}$. We examine the first-order case to gain some insight into the ideas behind the proof techniques for general systems. When $n=1$, the problems \eqref{eq:hard-lasso1} and \eqref{eq:hard-lasso2} are equivalent, and therefore, we only focus on \eqref{eq:hard-lasso2}.  After establishing the optimality conditions for these problems, we will examine two types of attack structures. An attack structure refers to the pattern of attack occurrences. In other words, it involves the distribution of each time instance at which a large disturbance vector is injected into the system. Namely, we inspect the structure of the set $\mathcal{K}$.

The first attack structure is a deterministic attack model for which the attacks occur at every $\Delta$ time period. For instance, if $\Delta=2$, the set $\calK $ could be $\{1,3,5,\dots,2k+1 \}$, meaning that an agent injects a disturbance vector into the system at every odd time instance. Later, we investigate a probabilistic attack structure where each attack may occur with probability $p$ at each time instance $i$, independent of the past periods. We first define the deterministic attack model, borrowed from \cite{han2021}.
    \begin{definition}[$\Delta$-spaced Attack Structure]
        Given a positive integer $\Delta > 2$, the disturbance sequence $\{ \bar{d}_i \}_{i=0}^{T-1}$ is said to be $\Delta$-spaced if for every $i \in \{0,1, \dots, T - \Delta - 1 \}$ such that $\bar{d}_i \not = 0$, we have $\bar{d}_j  = 0$, for all $j \in \{i+1, \dots, i+\Delta-1 \}$ and $\bar d_{i+\Delta} \not = 0$. In addition, for $i \in \{0, 1, \dots, \Delta-1\}$, we must have at least one non-zero disturbance vector, i.e. $\bar{d}_i \not = 0$.
    \end{definition}
    
    We will show that the convex formulation \eqref{eq:hard-lasso2} exactly recovers $\bar{A}$ in the case of $\Delta$-spaced disturbance sequence with $\Delta \geq 2$.
    \begin{proposition}
        Consider a first-order autonomous system with $\Delta$-spaced disturbance sequence with $\Delta \geq 2$. Then, the convex formulation \eqref{eq:hard-lasso2} (or equivalently \eqref{eq:hard-lasso1}) has the unique solution $\bar{A}$ as long as the sample complexity satisfies the inequality $T \geq \Delta + 1$. \label{prop: delta}
    \end{proposition}
    
    This proposition implies that whenever there are more than $\Delta+1$ data samples, the exact recovery is guaranteed to be achieved. Note that Proposition \ref{prop: delta} does not make any assumption on the vector set $\{\bar d_i: i \in \mathcal{K}\}$ and each element of the set could be arbitrarily large and correlated as long as they are finite. As a result, regardless of the severity of the attack, an exact recovery is guaranteed for \eqref{eq:hard-lasso1} and \eqref{eq:hard-lasso2}.  One important implication of Proposition \ref{prop: delta} is for the case where there is a $\Delta$-spaced disturbance sequence with $\Delta=2$, meaning that half of the observations are corrupted. In the robust regression estimation literature, exact recovery is possible only if the number of attacked observations is less than half of the total observations. The main difference between robust regression and system identification problems is that the observations are correlated with each other in the latter. This enables exact recovery for the convex formulation even if half of the data is corrupted via an adversarial agent. The proof of Proposition \ref{prop: delta} is based on the following lemma.
    \begin{lemma}(Theorem 1  in \cite{han2021})
     Consider the convex optimization problem \eqref{eq:hard-lasso2}. If $\sum_{i \not \in \mathcal{K}} |x_i| > \sum_{i \in \mathcal{K}} |x_i|$, then $\bar A$ is the unique solution to the problem.\label{lem: unique}
    \end{lemma}

    The proof of Lemma \ref{lem: unique} is based on the KKT conditions of the problem provided earlier. A natural question arises as to whether one can generalize the above result to higher-order systems. The next proposition extends Proposition \ref{prop: delta} to autonomous dynamical systems with an arbitrary order $n$ under a $\Delta$-spaced disturbance sequence with $\Delta \geq n+1$.
    \begin{proposition}
        Consider an autonomous system of order $n$ under a $\Delta$-spaced disturbance sequence with $\Delta \geq n+1$. Suppose that $\bar A $ is diagonalizable with eigenvalues, $\bar \lambda_l, l=1,2,\dots,n$, and that the condition
        \begin{align}
            \bar d_{i+\Delta} \in \mathrm{span}\{\bar d_{i}, \bar A \bar d_{i}, \dots, \bar A^{\Delta-2}\bar d_{i} \},\quad \forall i =0, 1, \dots, T-1 \label{eq: span}
        \end{align}
        is satisfied. Then, $\bar A$ is a solution to the convex formulation \eqref{eq:hard-lasso2} if $T \geq n + \Delta$, provided that
        \begin{align}
            \left| \sum_{k_1 + \dots + k_n = \Delta - n} \bar \lambda(k_1, \dots, k_n)  \right| \leq \sum_{t= 0}^{\Delta - n-1} \left| \sum_{k_1 + \dots + k_n = t} \bar \lambda(k_1, \dots, k_n)  \right|, \label{eq: ugly-eq}
        \end{align}
        where the notation $ \bar \lambda(k_1, \dots, k_n)$ denotes $\bar \lambda_1^{k_1} \times \bar \lambda_2^{k_2} \times \cdots \times \bar \lambda_n^{k_n}$.
        \label{prop: general-deter}
    \end{proposition}
    
\begin{figure}
    \centering
    \includegraphics[width=88mm]{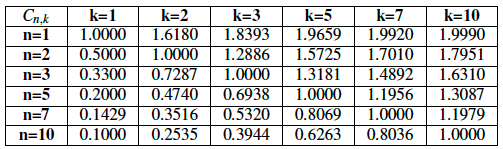}
    \caption{Upper-Bound Value $C_{n,k}$ for Different Values of $n$ and $k$.}
    \label{tb: table}
    \end{figure}

    This result is a generalization of Proposition \ref{prop: delta}, and we do not require all the eigenvalues of $\bar A$ to lie inside the unit circle (i.e., it allows the violation of Assumption 1). The condition \eqref{eq: span} is necessary to ensure that the KKT condition is satisfied, which eliminates the alignment of the attack vectors with eigenspaces of the matrix $\bar A$. In real-life applications, this circumstance can be avoided by injecting a small perturbation to the system. To gain insight into equation \eqref{eq: ugly-eq}, which involves the product of eigenvalues, consider a special case where $\bar A$ has the eigenvalue $\lambda$ with multiplicity $n$ and $n$ distinct eigenvectors. In this case, we can simplify \eqref{eq: ugly-eq} as follows. Define $k:= \Delta - n$. Then, \eqref{eq: ugly-eq} is equivalent to
    \[ {n + k -1 \choose k} |\lambda|^k - \sum_{i=0}^{k-1} {n + i -1 \choose i} |\lambda|^i < 0. \]
    This condition is satisfied if $|\lambda| \leq C_{n,k}$, where $C_{n,k}$ denotes the upper bound on the eigenvalue magnitudes given the parameters $n$ and $k$. Figure \ref{tb: table}  summarizes the values of $C_{n,k}$ for different choices of $n$ and $k$. Note that $C_{n,k} \leq C_{m,k}$ if $n >m$ and $C_{n,k} \leq C_{n,l}$ if $k < l$, due to the definition of $C_{n,k}$. It can be shown that $C_{1,k} \xrightarrow{} 2$ as $k \xrightarrow{} \infty$. As a result, $|\lambda| \leq C_{n,k} \leq C_{1,k} \xrightarrow{} 2$. This shows that the stability of the system is not necessary for exact recovery when the attack vectors are injected less frequently. In addition, whenever $k=n$ or $\Delta = 2n$, $|\lambda| < 1$ is sufficient for exact recovery as suggested by Proposition \ref{prop: general-deter}. This conclusion is analogous to the stability of the system.
    Proposition \ref{prop: general-deter} can still be applied to problem \eqref{eq:hard-lasso1}. However, the KKT conditions will differ due to the subdifferential of the $\ell_2$ and $\ell_1$ norms. In fact, they both have a similar shape. Therefore, one can show that this proposition still holds with the same condition even if convex formulation \eqref{eq:hard-lasso1} with the $\ell_1$ norm of the disturbance vectors is used.
    
    It is natural to ask whether it is possible to learn the system when there is more corrupted data than clean data. We cannot use a $\Delta$-spaced disturbance sequence model because the minimum value of $\Delta$ is $2$, which does not allow the size of corrupted data to exceed the size of clean data. Thus, we investigate a probabilistic attack structure. In this structure, a non-zero disturbance vector $\bar d_i$ is injected into the system at time instance $i$ with probability $p>0$, which is independent of the past and future time periods. To address this, we consider a probabilistic attack model where there is a parameter $p$ specifying the probability of an attack at each time instance. Specifically, given a time instance $i$, $\bar d_i$ is non-zero with probability $p$, and this is independent of all previous and future time instances. As a result, the event of having an attack at each time instance is identically and independently distributed with a Bernoulli distribution with parameter $p$. Nevertheless, the attack vectors are still allowed to be correlated with each other. Our goal is to discover the properties of \eqref{eq:hard-lasso1} and \eqref{eq:hard-lasso2} for an arbitrary value of $p$, especially $p > 0.5$. We make the following stealth attack assumption.
    \begin{assumption}\label{asp:stealth}
        For each $k\in\calK$, the attack vector is defined by
        \[ \bar{d}_k := \bar{\ell}_k \bar{f}_k,\quad \text{where } \bar{\ell}_k \in\mathbb{R}\text{ and }\bar{f}_k \in\mathbb{S}_2(1). \]
        where $\bar{f}_k $ plays the role of the direction of the attack while $\bar{\ell}_k$ plays the role of the length (that is allowed to take negative values too). Define the filtration
        \[ \calF_k := \sigma\{ x_1,\dots,x_k \},\quad \forall k\in\{0,\dots,T-1\}. \]
        For all $k\in\calK$, conditioning on $\calF_{k}$, the following statements hold:
        \begin{enumerate}
            \item $\bar{\ell}_k$ is independent from the direction $\bar{f}_k$;
            \item The direction $\bar{f}_k$ obeys the uniform distribution on $\mathbb{S}_2(1)$;
            \item $\bar{\ell}_k$ is mean-zero and sub-Gaussian with parameter ${\sigma}$;
            \item The variance of $\bar{\ell}_k$ is $\sigma_k^2 \in [c^2\sigma^2, {\sigma}^2]$ for some constant $c > 0$.
        \end{enumerate}
    \end{assumption}

    Under the stealth assumption, the length $\bar{\ell}_k$ can depend on the previous attacks $\bar{d}_{k'}$, and in particular $\bar{\ell}_{k'}$ and  $\bar{f}_{k'}$ for $k' < k$. In addition, we note that the above assumption of symmetry of the disturbance vectors reflected in $\bar{f}_k$ is not restrictive and corresponds to stealth attacks. If this assumption does not hold, the attacks may be detectable, and their effects could be nullified, or the system could be stopped to investigate the possible influence from outside agents. For an attack to be stealthy, its value should be zero on expectation, and our assumption has a similar flavor. If the symmetric assumption does not hold, it has been shown that there is a bias in estimation, and there is no way to avoid this bias \cite{chen2021bridging}. In the special cases when the length distribution is Gaussian or bounded, the constant $c$ is equal to $1$. Furthermore, we mention that the uniform distribution assumption of $\bar{f}_k$ can be relaxed to an arbitrary distribution on the sphere with zero mean and full-rank covariance matrix. In that more general case, the sample complexity in Theorems \ref{thm: exact-l2-general-no-input}-\ref{thm: general-l1} will depend on the conditional number of the covariance matrix, which is equal to $1$ under Assumption \ref{asp:stealth}.

Since the KKT conditions include random variables and random sets due to the randomness in the attack structure, it is not possible to obtain deterministic sample complexity for exact recovery as in Proposition \ref{prop: general-deter}. Therefore, it is essential to quantify the required number of samples for exact recovery with high probability using non-asymptotic analysis. Under Assumption \ref{asp:stealth}, the attack vector at time $i$, $\bar d_i$, has a sub-Gaussian distribution with parameter $\sigma$ given $\calF_{i}$, as described in Assumption \ref{asp:stealth}. The sub-Gaussianity assumption does not specify the distribution of the disturbance vector but assures that the disturbance vectors have light tails. For instance, any distribution over a bounded space is sub-Gaussian, making this assumption extremely mild. As a result, the sub-Gaussian assumption is not restrictive.

The KKT conditions for exact recovery, which are necessary and sufficient, can be restated as
\[ \exists \gamma_i \in \partial \| 0 \|_\circ, \quad \forall i \not \in \calK\quad \mathrm{s.t.}~ \sum_{i \not \in \mathcal{K}} x_i \otimes \gamma_i =
\sum_{i \in \mathcal{K}} x_i \otimes \partial \|\bar d_i \|_\circ.  \]
because of the properties of the subdifferentials at the origin. In order to simplify the analysis, we use the relationship between the unit balls of the $\ell_\infty$ and $\ell_2$ norms, that is $\frac{1}{\sqrt{n}} \mathbb{B}_\infty(1) \subseteq \mathbb{B}_2(1)$. Additionally, we examine the results for each coordinate of the subdifferentials since they are separable due to the properties of the $\ell_\infty$ norm. Therefore, the following propositions provide sufficient conditions to satisfy the KKT conditions.
\begin{proposition} 
The KKT conditions for the problem \eqref{eq:hard-lasso2} and \eqref{eq:hard-lasso1} are satisfied if there exist scalars $\gamma_i^l \in [-1,1], i \not \in \mathcal{K}, l = 1, \dots, n$ such that 
        \begin{align}
             \sum_{i \not \in \mathcal{K}} \gamma_i^l x_i/\sqrt{n}  =
\sum_{i \in \mathcal{K}}  \partial \| \bar d_i \|_2^l  x_i, \quad \forall l=1,\dots, n \label{eq: sufficient-2}
        \end{align}
and
\begin{align}
    \sum_{i \not \in \mathcal{K}} \gamma_i^l x_i  =
\sum_{i \in \mathcal{K}}  \partial \| \bar d_i \|_1^l  x_i, \quad \forall l=1,\dots, n, \label{eq: sufficient-1}
        \end{align}
respectively. Here, $\partial \| \bar d_i \|_\circ^l$ is the $l$-th element of the subgradient. \label{prop: equalitiy}
\end{proposition}

Because analyzing the conditions \eqref{eq: sufficient-2} and \eqref{eq: sufficient-1} directly is cumbersome, we investigate the equivalent condition provided in the lemma below, derived using Farkas' lemma \cite{Farkas1902} and the duality of linear programs.
\begin{lemma} Given a matrix $\mathbf{F}  \in \bR^{n \times m}$ and the vector $g \in \bR^n$, the following statements are equivalent:
\begin{itemize}
    \item[i)] There exists a vector $w \in \bR^m$ with $\|w\|_\infty \le 1$ satisfying $\mathbf{F} w = g$. 
    \item[ii)] For every $z \in \bR^{n}$ with $\| z \|_2 = 1$, it holds that $f(z) := z^T g + \| z^T \mathbf{F} \|_1 \ge 0$. 
\end{itemize} \label{lem: farkas}
\end{lemma}

It is important to notice that the conditions \eqref{eq: sufficient-2} and  \eqref{eq: sufficient-1} amount to finding a vector for the set of equations in the form of $\mathbf{F} w = g$ where $w$ is restricted as $\| w \|_\infty \le 1$. Given a coordinate $l$, the matrix $\mathbf{F}  \in \bR^{n \times (T-|\calK|)}$ associated with the conditions \eqref{eq: sufficient-2} and  \eqref{eq: sufficient-1} is a matrix with columns $\frac{x_i}{\sqrt{n}}$ and $x_i$, and the vector $g \in \bR^n$ is $\sum_{i \in \mathcal{K}} \partial \| \bar d_i \|_2^l  x_i$ and $ \sum_{i \in \mathcal{K}} \partial \| \bar d_i \|_1^l  x_i$, respectively. Moreover, the vector $w \in \bR^{T-|\calK|}$ has the elements $\gamma_i^l, i \not \in \calK$ for both conditions. Hence, we study the second statement in Lemma \ref{lem: farkas}. We use the union bound to study the satisfaction of this condition. However, there are infinitely many points inside the $\ell_2$ unit ball $\mathbb{B}_2(1)$. In order to show that the function $f(z)= z^T g + \| z^T \mathbf{F}  \|_1$ is non-negative at every point inside the $\ell_2$ unit ball, we employ the discretization technique that uses a finite set of points. The set of such points is called the cover of the unit ball.
\begin{definition}[Covering Number \cite{wainwright_2019}]
    Let $(\mathbb{T}, \rho)$ be a compact metric space with a set $\mathbb{T}$ and a norm operator $\rho$. $\epsilon$-cover of the set $\mathbb{T}$ with respect to the norm $\rho$ is a set $\{\theta^1, \theta^2, \dots, \theta^N\} \subset \mathbb{T}$ such that for each $\theta \in \mathbb{T}$, there exists some $i \in \{1, \dots, N\}$ such that $\rho(\theta, \theta^i) \le \epsilon$. The $\epsilon$-covering number $\calN(\epsilon, \mathbb{T}, \rho)$ is the cardinality of the smallest $\epsilon$-cover.
\end{definition}

Given a $\epsilon > 0$, the logarithm of the covering number of the unit ball or the metric entropy of the unit ball can be upper bounded using the volumetric arguments of the balls. Indeed, the number of $\epsilon$ balls exceeding $ \exp \{n \log \left( 1 + 2/\epsilon\right)\}$ is sufficient to cover the unit ball with balls of radius $\epsilon$.
\begin{lemma}[Covering Number of the Unit Ball \cite{wainwright_2019}]
    Given an $n$-dimensional unit ball $\mathbb{B}(1)$ with the norm $\| \cdot\|$,
    \[ \mathbb{B}(1) = \{ x \in \bR^n : \| x \| \le 1 \}, \]
    the logarithm of the $\epsilon$-covering number, i.e., the metric entropy of the unit ball, can be upper bounded by
    \[ \log \calN(\epsilon, \mathbb{B}(1), \| \cdot \|) \le n \log \left( 1 + \frac{2}{\epsilon} \right). \] \label{lem: covering}
\end{lemma}

We show that the function $f(z)$ can be lower bounded by some positive number $\theta > 0$ at every point in the $\epsilon$-cover of the unit circle with high probability, and that the function value inside the $\epsilon$-ball does not change more than this positive number $\theta$ with high probability. Thus, $f(z)$ must be non-negative at every point of the unit circle with high probability.  Utilizing this idea, the next theorem shows that the required number of samples for the exact recovery grows with $n^2$ and $(1-p)^{-2}$ for the general systems of order $n$.
\begin{theorem}
Consider an autonomous system of order $n$ under a probabilistic attack model with frequency $p$.
Suppose that Assumptions \ref{asp:stable} and \ref{asp:stealth} hold.
Then, for all $\delta\in(0,1]$, if the time horizon satisfies $T \geq \Theta(T_{\text{sample}})$, where $T_{\text{sample}}$ is defined as 
\begin{align*}
 n R \left[ n\log(nR) + \log\left(\frac{1}{\delta}\right) \right],
\end{align*}
and
        \begin{align*}
            R := \max \Bigg\{ & \frac{\log(1/c)}{nc^4 p(1-p) \log(1/\rho)}, \\ 
            & \frac{\log^2(1/c)}{c^{10} (1-p)^2 (1-\rho)^3\log^2(1/\rho)}, \frac{1}{np(1-p)} \Bigg\},
        \end{align*}
with $\rho$ denoting the largest magnitude of the eigenvalues of $\bar{A}$, then $\bar A$ is a solution to the convex optimization \eqref{eq:hard-lasso2} with probability at least $1-\delta$. \label{thm: exact-l2-general-no-input}
\end{theorem}

An implication of the above theorem is that even when $p$ is large (e.g., $p > 0.5$) corresponding to the system being under attack frequently, exact recovery of the system dynamics is still possible as long as the time horizon is above the threshold. 
Similar results can be obtained if one prefers to use problem \eqref{eq:hard-lasso1} to recover the system matrix $\bar A$.
\begin{theorem}
 Under the same assumptions as in Theorem \ref{thm: exact-l2-general-no-input}, if the time horizon $T$ satisfies $T \geq \Theta(T_{\text{sample}})$, where $T_{\text{sample}}$ is defined as  
\begin{align*}
    R \left[ n\log(nR) + \log\left(\frac{1}{\delta}\right) \right],
\end{align*}
and $R$ is defined in Theorem \ref{thm: exact-l2-general-no-input}, then $\bar A$ is a solution to the convex optimization \eqref{eq:hard-lasso1} with probability at least $1-\delta$.  \label{thm: exact-l1-general-no-input}
\end{theorem}

The proof of Theorem \ref{thm: exact-l1-general-no-input} is highly similar to that of Theorem \ref{thm: exact-l2-general-no-input} and therefore, it is omitted. Because the conditions \eqref{eq: sufficient-2} and \eqref{eq: sufficient-1} differ by a factor of $\sqrt{n}$, the sample complexity results in those theorems differ by a factor of $n$.



The required amount of data increases with the value $(1-p)^{-2}$ and the order of the system $n$. Hence, as $p$ and $n$ increase, the number of samples for exact recovery with high probability grows. The results on sample complexity are intuitive: as the probability of having an attack increases, a larger time horizon is required for exact recovery. We note that the dependence on $p^{-1}(1-p)^{-1}$ is an artifact of the high probability bound. More specifically, this dependence guarantees that the number of attacks is bounded by $\Theta(pT)$ with high probability. 
In addition, if the system is at the verge of instability with eigenvalues close to the unit circle, the sample complexity increases significantly.  Even in the case when the probability $p$ is close to $1$, resulting in significantly more corrupt data than clean data, this result guarantees asymptotic exact recovery as long as there are a sufficient number of clean samples. 

Last but not at least, due to the logarithmic probability bound and the Borel-Cantelli lemma, Theorems \ref{thm: exact-l2-general-no-input} and \ref{thm: exact-l1-general-no-input} imply almost sure asymptotic convergence as a corollary. Almost sure convergence of random variables 
implies the convergence in probability and convergence in distribution for a sequence of random variables. Almost sure convergence of random variables is defined as below for completeness. 
 \begin{definition}[Almost Sure Convergence]
          A sequence of random variables $X_1, X_2, X_3, \dots$ converges to $X$ almost surely if
           \[ \mathbb{P} \left( \lim_{n \xrightarrow{} \infty} X_n = X \right) \rightarrow 1. \]
\end{definition}

The following corollary states that the sequence of estimators over time converges to the true system matrices almost surely. 
\begin{corollary}
Under the same assumptions as in Theorem (\ref{thm: exact-l2-general-no-input}), $\bar A$ is almost surely a solution of convex formulations \eqref{eq:hard-lasso2} and \eqref{eq:hard-lasso1} when $T$ goes to infinity. 
\end{corollary}

\section{Systems with Input Sequence}

It is desirable to understand the role of an input sequence in exact recovery because the majority of dynamical systems are controlled by an external input. Since the input sequence is generated by a controller, one can design it in such a way that it accelerates the exact recovery. In the non-autonomous case, the system dynamics is given as $x_{i+1} = \bar A x_i+ \bar B u_i + \bar d_i, i = 0, \dots, T-1$, where $\bar A \in \mathbb{R}^{n \times n}$ and $\bar B \in \mathbb{R}^{n \times m}$. Similar to the autonomous case, the true system matrices $\bar A$ and $\bar B$ are not known and the goal is to obtain these matrices using the state trajectories and the sequence of inputs. Unlike the disturbance vectors $\bar d_i, i\in \{0, \ldots, T-1\}$, the sequence of system states $x_i, i\in \{0, \ldots, T\}$, and the sequence of the inputs $u_i, i\in \{0, \ldots, T-1\}$ are known. We will investigate the estimators \eqref{eq:lasso-2} and \eqref{eq:lasso-1} defined earlier.

We choose the input vectors $u_i$ to be Gaussian given $\calF_{i}$. This allows us to obtain a high-probability bound for the exact recovery of the matrices $\bar A$ and $\bar B$. A random input sequence is commonly used in system identification and online learning because it enables the exploration of the system to learn the system dynamics faster. The Gaussian input assumption may seem restrictive. Nevertheless, it is satisfied when $u_i$ is designed in the linear feedback form as $u_i = Kx_i + \omega$. Conditioning on $\calF_i$, if the input is excited with Gaussian noise $\omega$, the input vector $u_i$ is also Gaussian. Therefore, the most common input sequence used in optimal control satisfies this assumption. Note that the closed loop system could be written as $x_{i+1} = (\bar A + \bar B K) x_i + \bar B \omega + \bar d_i$. Thus, the problem is equivalent to estimating the matrices $(\bar A + \bar B K)$ and $\bar B$ when the linear feedback control is used. 

The KKT conditions for the exact recovery that are both necessary and sufficient can be restated as 
    \begin{align*}
     \exists \gamma_i \in \partial \| 0 \|_\circ\quad \forall~i \not \in \calK \text{ s.t. } \sum_{i \not \in \mathcal{K}} x_i \otimes \gamma_i =
    \sum_{i \in \mathcal{K}} x_i \otimes \partial \|\bar d_i \|_\circ
    \end{align*}
    and 
    \begin{align*}
     \exists \mu_i \in \partial \| 0 \|_\circ\quad \forall~i \not \in \calK \text{ s.t. } \sum_{i \not \in \mathcal{K}} u_i \otimes \mu_i =
    \sum_{i \in \mathcal{K}} u_i \otimes \partial \|\bar d_i \|_\circ.
    \end{align*}
The first set of conditions corresponds to the KKT conditions for the system states while the second set is for the KKT conditions for the input sequence. Similar to Proposition \ref{prop: equalitiy}, the sufficient conditions can be tightened so that the equations become coordinate-wise separable. 
\begin{proposition} 
The KKT conditions for problem \eqref{eq:lasso-2} are satisfied if there exist scalars $\gamma_i^l, \mu_i^l \in [-1,1]$ for all $i \not \in \mathcal{K}, l \in\{ 1, \dots, n\}$ such that 
    \begin{align}
             \sum_{i \not \in \mathcal{K}} \gamma_i^l x_i/\sqrt{n}  =
\sum_{i \in \mathcal{K}}  \partial \| \bar d_i \|_2^l  x_i, \quad \forall l=1,\dots, n,                     \label{eq: general-suff-state}
        \end{align}
and
        \begin{align}
             \sum_{i \not \in \mathcal{K}} \mu_i^l u_i/\sqrt{n}  =
\sum_{i \in \mathcal{K}}  \partial \| \bar d_i \|_2^l  u_i, \quad \forall l=1,\dots, n, \label{eq: general-suff-input}
        \end{align}
         where $\partial \| \bar d_i \|_2^l$ denotes the $l$-th element of the subgradient.
\label{prop: general}
\end{proposition}

The proof of Proposition \ref{prop: general} is omitted because it relies on the same technique as in Proposition \ref{prop: equalitiy}. As in the case of autonomous systems, two sets of equations that guarantee the satisfaction of the KKT conditions can be written for problem \eqref{eq:lasso-1} by omitting the factor $\sqrt{n}$. To establish the exact recovery guarantees, we require the following controllability assumption.
\begin{assumption}\label{asp:controllability}
The ground truth $(\bar{A}, \bar{B})$ satisfies
\begin{align*}
    \mathrm{rank}\left\{ \begin{bmatrix}
            \bar{B} & \bar{A}\bar{B} &\cdots & \bar{A}^{n-1}\bar{B}
        \end{bmatrix} \right\} = n.
\end{align*}
\end{assumption}

Intuitively, the controllability of a non-autonomous system denotes the ability to move a system around in its entire state space using the admissible manipulations, namely, the input sequence $\{u_t\}_{t=0}^{T-1}$. Controllability is an important property of a control system and plays a crucial role in many control problems, such as stabilization of unstable systems by feedback.
Under the above assumption, we implement the non-asymptotic analysis of the general non-autonomous system in a similar fashion to Theorem \ref{thm: exact-l2-general-no-input} using the covering arguments and Farkas' lemma.
\begin{theorem}
        Consider an autonomous system of order $n$ under a probabilistic attack model with frequency $p$. Suppose that Assumptions \ref{asp:stable}, \ref{asp:controllability} and the first three conditions in Assumption \ref{asp:stealth} hold. Assume also that the input vectors $u_i| \calF_{i}$ are selected to be independent from the attack vectors and obey the Gaussian distribution $\mathcal{N}(0, \frac{\xi^2}{m} I_m)$. 
        For all $\delta\in(0,1]$, let 
        %
        \begin{align*}
            T_{\text{sample}}^1 := n R_1 \left[ n\log(nR_1) + \log\left(\frac{1}{\delta}\right) \right]
        \end{align*}
        and 
        \begin{align*}
            T_{\text{sample}}^2 := n R_2 \left[ m\log(nR_2) + \log\left(\frac{1}{\delta}\right) \right],
        \end{align*}
        %
        where
        \begin{align*}
        & R_1 := \max \Bigg\{   \frac{\log(\kappa/c)}{nc^4  \log(1/\rho)},\frac{p \kappa^2}{c^{10} (1 - p)^2 (1-\rho)^2}, \\
         & \hspace{5em} \frac{p \kappa^2\log^2(\kappa/c)}{c^{10}(1-\rho)^2 \log^2(1/\rho) },\frac{1}{np} \Bigg\} \\
        & R_2 := \max\Bigg\{\frac{1}{np},\frac{p}{(1-p)^2}, \frac{m}{n} \Bigg\}.
        \end{align*}
        Here, constants $c \in (0,1]$ and $\kappa\geq (1-\rho)^{-1}$ depend on $m$, $n$, $\sigma$, $\xi$ and $\bar{B}$. If the time horizon satisfies the inequality $T \geq \Theta[\max\{T_{\text{sample}}^1,T_{\text{sample}}^2\}]$, then $(\bar A, \bar B)$ is a solution to \eqref{eq:lasso-2} with probability at least $1-\delta$.
        \label{thm: general-l2}
    \end{theorem}
    
    We have obtained a high probability bound for the exact recovery of the system matrices $\bar A$ and $\bar B$. The first term in the sample complexity corresponds to the satisfaction of the KKT conditions for the state measurements $\{x_i\}_{i=0}^T$, whereas the second term corresponds to the satisfaction of the KKT conditions for the input sequence $\{u_i\}_{i=0}^{T-1}$. Similar to the case of autonomous systems, the sample complexity increases as the probability of disturbances increases. Because there is a logarithmic dependence on the satisfaction of the probability bound, Theorem \ref{thm: general-l2} and the application of the Borel-Cantelli lemma imply almost sure asymptotic convergence to the correct matrices $\bar A$ and $\bar B$. The sample complexity $T_{\text{sample}}^2$ is needed to satisfy the KKT conditions associated with on the input sequence. Compared with the previous theorems for the autonomous case, we require a sample complexity that scales with $p/(1-p)^2$ and terms depending on the spectral norm of $\bar A$. The introduction of the input sequence removes the requirement on the variance of the attack vectors. In addition, the dependence of the sample complexity on $p$ is improved from $1/(1-p)^{2}$ to $p/(1-p)^{2}$. Moreover, the dependence on the spectrum of $\bar{A}$ is reduced from $1/[(1-\rho)^{3}\log^2(1/\rho)]$ to $1/[(1-\rho)^{2}\log^2(1/\rho)]$. Finally, we mention that the dependence on $1/(np)$ is also to guarantee that the number of attacks is bounded by $\Theta(pT)$ with high probability.
    
    The following theorem studies problem \eqref{eq:lasso-1}.
    \begin{theorem}\label{thm: general-l1}
        Under the assumptions of Theorem \ref{thm: general-l2}, for all $\delta\in(0,1]$, let $T_{\text{sample}}^1$ and $T_{\text{sample}}^2$ be defined as
        \begin{align*}
        R_1 \left[ n\log(nR_1) + \log\left(\frac{1}{\delta}\right) \right]  \text{ and }R_2 \left[ m\log(nR_2) + \log\left(\frac{1}{\delta}\right) \right],
        \end{align*}
        %
        %
        where $R_1$ and $R_2$ are given in Theorem \ref{thm: general-l2}. If the time horizon satisfies the inequality $T \geq \Theta[\max\{T_{\text{sample}}^1,T_{\text{sample}}^2\}]$, then $(\bar A, \bar B)$ is a solution to \eqref{eq:lasso-1} with probability at least $1-\delta$.
    \end{theorem}

    As expected, even if more than half of the data are corrupted, that is $p>1/2$, the exact recovery is still attainable with high probability. We note that when the input sequence $u_i = Kx_i$ is used to control the system, this input sequence satisfies the assumptions in the above theorems if $x_i$ are sub-Gaussian. The closed-loop system with the matrix $(\bar A+\bar B K)$ results in a second solution $\hat A = \bar A+\bar B K$ and $\hat B = 0$. Nevertheless, the ground-truth system matrix pair $(\bar A, \bar B)$ is also a solution to our estimators. This phenomenon occurs due to the existence of multiple optimal solutions and it could be avoided if the input is excited with a small noise in the form of $u_i = Kx_i+\omega$. Moreover, if all the input vectors $u_i$ are set to zero, it is not possible to uniquely recover the system matrix $\bar B$. Nevertheless, because the input sequence $\{u_i\}_{i=0}^{T-1}$ is zero, the KKT conditions are trivially satisfied. Therefore, the estimators have multiple optimum solutions where $\bar B$ and $0$ matrices are possible solutions among all optimum solutions. 

   \section{Numerical Experiment}
    \begin{figure*}[]
    \centering
    \includegraphics[width=181mm]{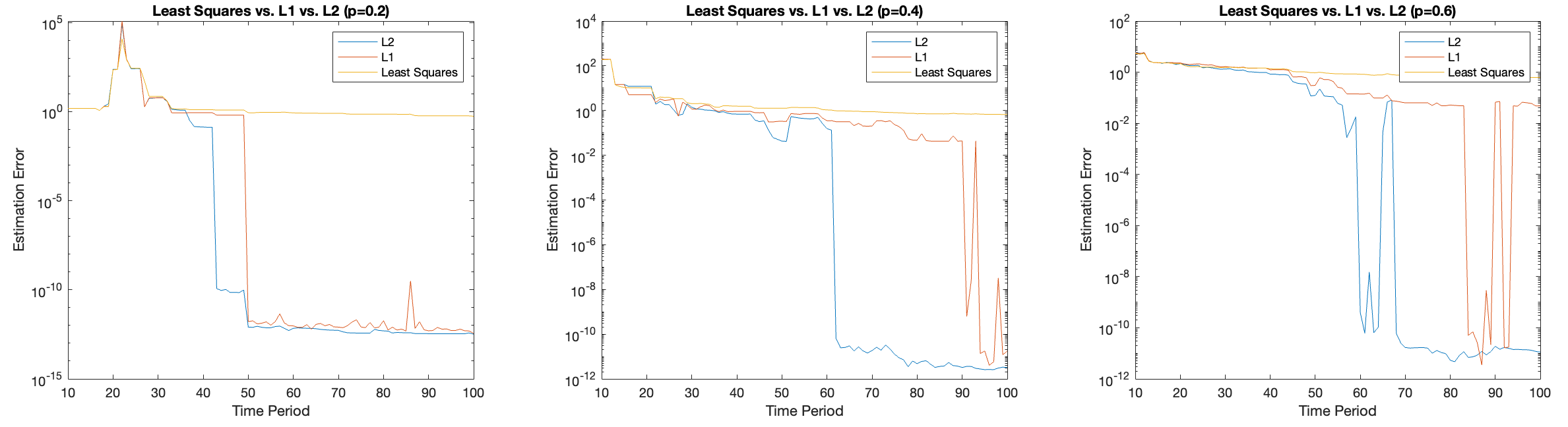}
    \caption{Estimation errors for Least-Squares, \eqref{eq:lasso-2}, and \eqref{eq:lasso-1} with attack probability of $p=0.2, 0.4, 0.6$ (left-to-right).}
    \label{fig:insulin}
    \end{figure*}
    \begin{figure}[]
        \centering
        \includegraphics[width=60mm]{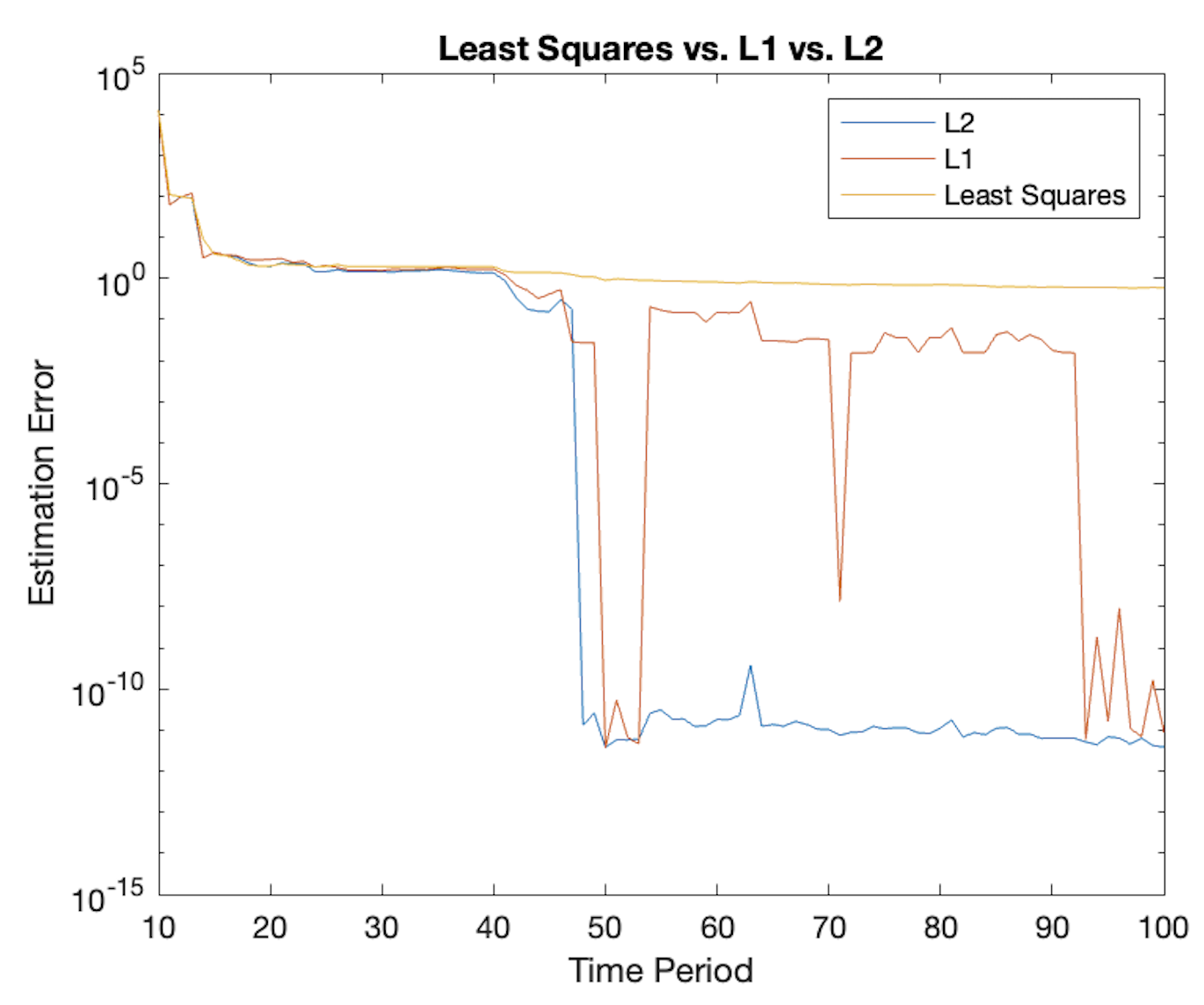}\\
        \includegraphics[width=60mm]{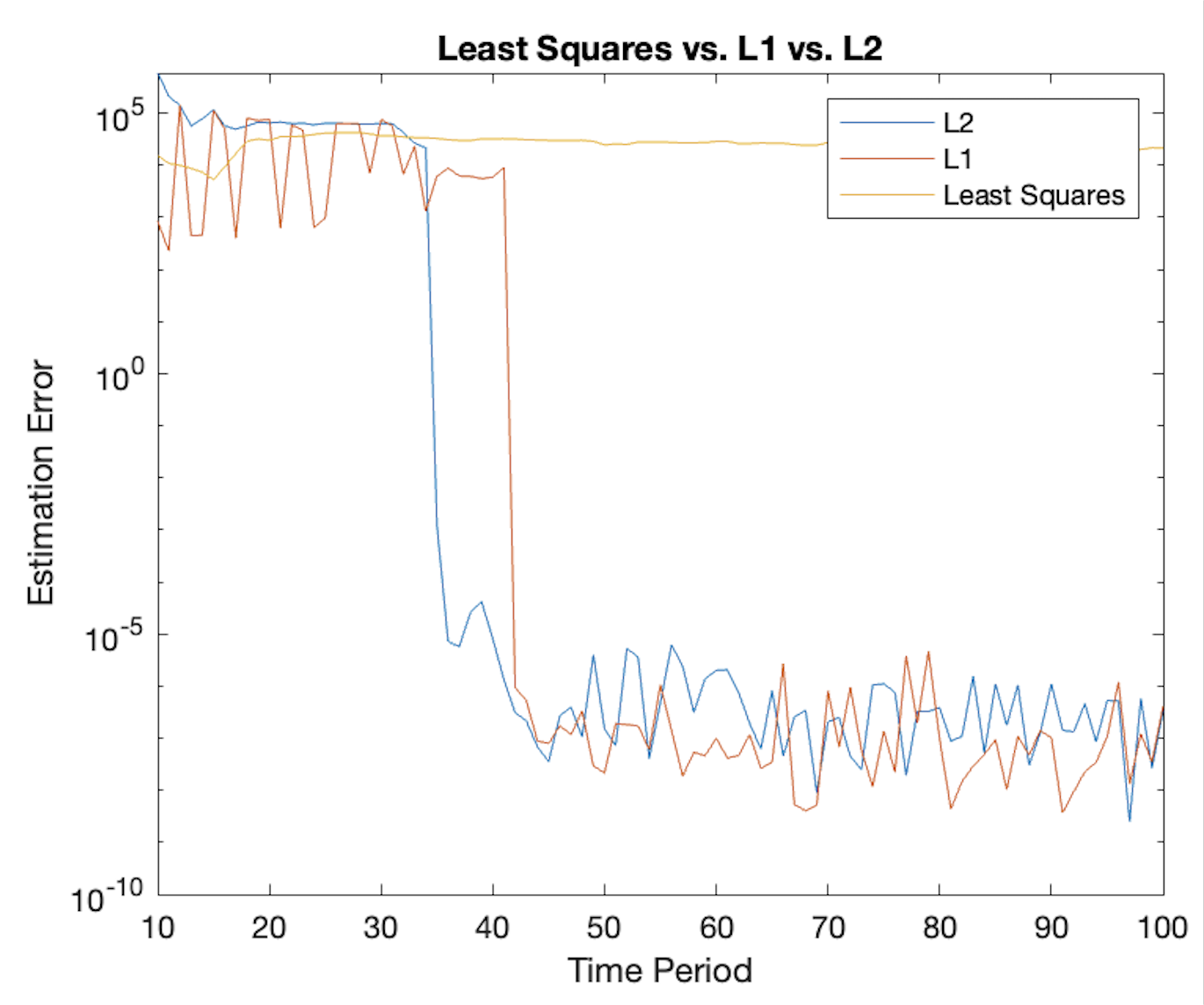}
        \caption{Estimation errors for Least-Squares, \eqref{eq:lasso-2}, and \eqref{eq:lasso-1} with attack probability $p=0.6$  not Sparse $d$ (top)  Sparse $d$  (bottom).}
        \label{fig:insulin2}
    \end{figure}
    We conduct a numerical experiment inspired by biomedical applications to demonstrate the results of this paper. We consider a compartmental model of blood sugar and insulin dynamics in the human body, as described in \cite{hovorka2002partitioning}. Accurately estimating the parameters of the dynamics is crucial when regulating the blood sugar level through the injection of a bolus of insulin into the system. Due to the complex structure of the human body, the dynamics vary among individuals. We consider a linear system based on Hovarka's model as follows \cite{compartment2018}:
    \begin{align*}
        \dot{x}_1 & = -k_{a1} x_1 - k_{b1} I + d_1,\\
        \dot{x}_2 & = -k_{a2} x_1 - k_{b2} I + d_2, \\
        \dot{x}_3 & = -k_{a3} x_1 - k_{b3} I + d_3, \\
        \dot{S}_1 & = -S_1/t_{max,I} + d_4, \\
        \dot{S}_2 & = S_1/t_{max,I} - S_2/t_{max,I} + d_5, \\
        \dot{I} & = S_2/(t_{max,I}V_I) - k_e I + d_6,
    \end{align*}
    where given a time-dependent variable $z(t)$, $\dot{z}(t)$ represents its derivative with respect to time $t$. The states $x_1, x_2, x_3$ represent the influence of insulin on the system of the body. $S_1$ and $S_2$ represent the absorption rate of insulin in the, directly and indirectly, accessible compartment models, respectively.  Lastly, the state $I$ represents the blood sugar level in the body. The disturbance $d_4$ corresponds to the bolus injection into the body, while the remaining disturbance vectors model sudden changes in the body due to diseases such as diabetes. Although the injected insulin amount could be known, the exact amount of insulin and its timing reaching the effective body parts are unknown. Hence, the $d_i$ values are treated as unknown. Even though the disturbance in this application is not a malicious attack, it exhibits similar characteristics for identification purposes: the arrival time of the bolus is unknown, and once it arrives, it has a large magnitude.

    In this experiment, we discretize the continuous-time system to obtain an LTI system using $\Delta_t = 0.5$. The resulting matrix $\bar A$ is stable.
    Our objective is to estimate the parameters $(k_{ai}, k_{bi}, t_{\text{max}, I}, V_I, k_e)$, where the true values are obtained from Table 1 in \cite{hovorka2004nonlinear}. We model the attack vectors given the historical data as zero-mean Gaussian random vectors with an identity covariance matrix with variance $10$. Thus, the attack vectors are conditionally independent, although they are dependent. We run our model with the probability of an attack being $p=0.2$, $p=0.4$, and $p=0.6$. We report the estimation error $| \hat A - \bar A|_F$ for the least-squares estimator, problem \eqref{eq:lasso-2}, and problem \eqref{eq:lasso-1}.
    
    Figure \ref{fig:insulin} suggests that our proposed estimators attain exact recovery while the least-squares estimator fails to do so. As the probability of having an attack $p$ increases, the number of required time periods for exact recovery grows proportionally to $p/(1-p)^2$. Note that there are more corrupted data than clean data in the case of $p=0.6$. Additionally, because there is no sparsity assumption on the attack vectors, \eqref{eq:lasso-2} performs slightly better than \eqref{eq:lasso-1}.
    
    We compare the performance of \eqref{eq:lasso-2} and \eqref{eq:lasso-1} by running a similar experiment with and without sparse disturbances. When the disturbances are sparse, $d_1, d_2, d_3, d_5$ are set to zero while $d_4$ and $d_6$ have the same Gaussian distribution as before. Figure \ref{fig:insulin2} shows that the two methods perform similarly when the attack vectors are also sparse.
    
    \section{Discussion and Conclusion}

We investigated the problem of learning LTI systems under adversarial attacks by studying two lasso-type estimators. We considered both deterministic and probabilistic attack models regarding the time occurrence of the attack and developed strong conditions for the exact recovery of the system dynamics. When the attacks occur deterministically every $\Delta$ period, exact recovery is possible after $n+\Delta$ time steps. Moreover, if the system is attacked at each time instance with probability $p$, the system matrices are recovered with high probability when $T$ is on the order of $\Theta((1-p)^{-2})$ and a polynomial in the dimension of the problem. Similar results were obtained when the system is controlled by an input sequence. These findings were supported by a numerical experiment in biology that to validate the non-asymptotic analytic results. This work provides the first set of mathematical guarantees for the robust non-asymptotic analysis of dynamic systems.

Since our estimators have non-smooth objective functions, closed-form solutions to the optimization problem are not obtainable. We did not provide any specific numerical algorithm to solve the provided estimation problems. However, both \eqref{eq:lasso-2} and \eqref{eq:lasso-1} are convex optimization problems, allowing the use of the subgradient descent algorithm to obtain these estimators. It is a well-established result that the subgradient algorithm has a convergence rate on the order of $\frac{1}{\sqrt{k}}$, where $k$ is the iteration update number. Although the algorithm is considered fast, one possible extension of this work would be to design an algorithm to predict and update $(\hat A_{t+1}, \hat B_{t+1})$ using the latest estimation $(\hat A_t, \hat B_t)$ and the new data $(x_{t+1}, u_t)$, instead of solving the problem from scratch at each time period. Initial experiments hinted that a single subgradient update at each iteration using the new information, $(x_{t+1}, u_t)$, asymptotically converges to the true system matrices. We leave the analysis of this algorithm and online control of dynamic systems under adversaries as future work.

\section*{Appendix}

\subsection{Proofs for Results in Main Part}

\subsubsection{Proof of Proposition \ref{prop: delta}}

    The proof of Proposition \ref{prop: delta} is established based on Lemma \ref{lem: unique} defined in the paper. Let $i_1, i_2, \dots$ be the set of attack times over time horizon $T$. Therefore, $\mathcal{K} = \{ i_1, i_2, \dots\}$. Due to $\Delta$-spaced attack model, the first attack time must be smaller than $\Delta$, i.e., $i_1 \leq \Delta$. Since $x_0 = 0$, we have $x_t = 0$ for $t = 0, 1, \dots, i_1$. Define $\mathbb{N}$ as the set of natural numbers. We can utilize Lemma \ref{lem: unique} to show that $\bar A$ is the unique solution. Using these facts, we can decompose the sum of the magnitudes of the states at non-attack times as
        \begin{align*}
            \sum_{i \not \in \mathcal{K}} |x_i| & = \sum_{i \not \in \mathcal{K}, \; i> i_1} |x_i|   = \sum_{i \in \mathcal{K}'} |x_i| + \sum_{i \in \mathcal{K}''} |x_i|,
        \end{align*}
        where $\mathcal{K}^{c+} = \mathbb{N} \backslash ( \mathcal{K} \cup \{ 0, 1, \dots, i_1 - 1 \}$), $\mathcal{K}' = \mathcal{K}^{c+} \backslash \mathcal{K}''$, and $\mathcal{K}'' = \{ i_2 -1 , i_3 -1 , \dots \}$. The second term on the right-hand side is the sum of magnitudes at the time step just before the attack while the first term covers the rest of the magnitudes of the states. In addition, the magnitudes of the states at attack times can be written as
            \begin{align*}
            \sum_{i \in \mathcal{K}} |x_i| & = \sum_{i \in \mathcal{K}, \; i \geq i_2} |x_i|  = \sum_{i \in \mathcal{K}''} |\bar A x_i| = \sum_{i \in \mathcal{K}''} |\bar A| | x_i|.
        \end{align*}
        The second equality follows from the fact that $x_{i_k} = \bar A x_{i_{k}-1}$ due to lack of attack. We compare the sum of the magnitudes of the states at attack times for the non-attack times to check if the condition in Lemma  \ref{lem: unique} holds: 
        \begin{align}
            \sum_{i \not \in \mathcal{K}}|x_i| - \sum_{i \in \mathcal{K}}|x_i| & = \sum_{i \in \mathcal{K}'} |x_i| + \sum_{i \in \mathcal{K}''} |x_i| - \sum_{i \in \mathcal{K}''} |\bar A| | x_i| \notag \\
            & = \sum_{i \in \mathcal{K}'} |x_i| + (1-|\bar A|) \sum_{i \in \mathcal{K}''} | x_i| > 0. \label{eq: proof-line}
        \end{align}
        Note that the term $\sum_{i \not \in \mathcal{K}} |x_i|$ becomes positive at time period $i_1 + 1$ while $\sum_{i \in \mathcal{K}} |x_i|$ is positive first time at time step $i_2$. Consequently, the strict inequality for \eqref{eq: proof-line} holds for every time step after $i_1$ because $(1-|\bar A|)>0$ by assumption. As a result, we have a unique and exact recovery for every time period $T \geq \Delta + 1  \geq i_1 + 1$.

\subsubsection{Proof of Proposition \ref{prop: general-deter}}

        By using \eqref{eq: kkt2}, the necessary and sufficient condition for this problem is 
        \[ 0 \in \sum_{i \not \in \mathcal{K}} x_i \otimes \partial \|(\bar A - A)x_i \|_2 + \sum_{i \in \mathcal{K}} x_i \otimes \partial \|(\bar A - A)x_i + \bar d_i\|_2. \]
        Then, $\bar A$ is a solution to the problem if and only if 
        \begin{align}
            0 \in \sum_{i \not \in \mathcal{K}} x_i \otimes \partial \|0\|_2 + \sum_{i \in \mathcal{K}} x_i \otimes \partial \| \bar d_i\|_2. \label{eq:kkt}
        \end{align}
        Let $i_1$ be the time stamp of the first attack time. Then, we have $i_1 \in \{1, \dots, \Delta \}$. The set of attack times is $\mathcal{K} = \{ i_1, i_1 + \Delta, i_1 + 2\Delta, i_1 + 3\Delta, \dots \}$. Since $x_0 = 0$, we have $x_t = 0$ whenever $t = 0, 1, \dots, i_1$ and $x_{i_1 + 1} = \bar d_{i_1}$. Let $T = \Delta + i_1$, i.e., the time step at which a cycle of disturbance is completed. In this case, the sufficient condition \eqref{eq: kkt2} can be written as 
        \begin{align*}
            0 & \in \sum_{t=1}^{\Delta-1} x_{i_1 + t} \otimes \partial \|0\|_2 + x_{i_1 + \Delta} \otimes \partial \|\bar d_{i_1 + \Delta}\|_2 \\
            & = \sum_{t=0}^{\Delta-2} \bar A^t \bar d_{i_1} \otimes \partial \|0\|_2 +   \bar A^{\Delta-1} \bar d_{i_1} \otimes \frac{\bar d_{i_1+\Delta}}{\|\bar d_{i_1+\Delta}\|}.
        \end{align*}
        The matrix $0$ may belong to the right-hand side term for arbitrary $\bar d_{i_1+\Delta}$ if $\bar d_{i_1+\Delta} \in span\{\bar d_{i_1}, \bar A \bar d_{i_1}, \dots, \bar A^{\Delta-2}\bar d_{i_1} \}$. This is satisfied by the assumption in the proposition statement.
        
        However, this is not sufficient to ensure that KKT condition \eqref{eq: kkt2} holds. The reason is that $\partial \|0\|_2 = \{ x \in \mathbb{R}^n : \| x \|_2 \leq 1\}$. The vectors chosen for $\partial \|0\|_2$ have a bounded norm. Therefore, we need a condition that bounds the norm of the columns of $\bar A^{\Delta-1} \bar d_{i_1} \otimes \frac{\bar d_{i_1+\Delta}}{\|\bar d_{i_1+\Delta}\|_2}$, so it can be expressed as a linear combination of the vectors $\{\bar d_{i_1}, \bar A \bar d_{i_1}, \dots, \bar A^{\Delta-2}\bar d_{i_1} \} $.
        Let $(\lambda_j, v_j)$ be eigenvalue-eigenvector pairs for the matrix $\bar A^T$. Let $e_1, \dots, e_{\Delta-1} \in \partial \|0\|_2$. Then, the KKT condition can be written as follows after dropping the sub-index ${i_1}$:
        \begin{align*}
            0 \in e_1 \bar d^T + e_2 \bar d^T \bar A^T + \dots + e_{\Delta-1}\bar d^T (\bar A^T)^{\Delta - 2} + f \bar d^T (\bar A^T)^{\Delta-1},
        \end{align*}
               where $f = \frac{\bar d_{i_1+\Delta}}{\|\bar d_{i_1+\Delta}\|_2}$ and $\|f\|_2 = 1$.
        If we multiply the equation above by the eigenvector $v_j$ of $\bar A^T$, we obtain
        \begin{align*}
            0 & \in e_1 \bar d^T v_j + \dots + e_{\Delta-1}\bar d^T (\bar A^T)^{\Delta - 2}v_j + f \bar d^T (\bar A^T)^{\Delta-1}v_j \\
            & \in (e_1 + \lambda_j e_2 + \dots + \lambda_j^{\Delta-2} e_{\Delta-1} + \lambda_j^{\Delta-1} f)\bar d^T v_j.
        \end{align*}
        Note that because $\bar{A}$ is diagonalizable, we only need to satisfy this condition along the direction of each eigenvector, since all eigenvectors span the whole space. Therefore, the KKT condition holds if 
        \begin{align*}
            0 & \in e_1 + \lambda_j e_2 + \dots + \lambda_j^{\Delta-2} e_{\Delta-1} + \lambda_j^{\Delta-1} f, \quad \forall j = 1, \dots, n.
        \end{align*}
        There are $(\Delta-1)n$ free variables and $n^2$ equations. One can use the substitution to eliminate $n^2$ variables, which leads to
        \begin{align*}
            \sum_{k_1 + \dots + k_n =  \Delta - n} & \lambda(k_1,\dots, k_n) f  = \\
            & \sum_{t=0}^{\Delta-n-2} \sum_{k_1 + \dots + k_n =t} \lambda(k_1,\dots, k_n)  e_{t+n+1}.
        \end{align*}
        Taking the norm of both sides and using the triangle inequality yields that
        \begin{align*}
            & \left|\sum_{k_1 + \dots + k_n = \Delta - n} \lambda(k_1,\dots, k_n) \right|  \|f\|_2 
            \\ & \leq \sum_{t=0}^{\Delta-n-1} \left| \sum_{k_1 + \dots + k_n =t} \lambda(k_1,\dots, k_n)\right|  \| e_{t+n+1}\|_2.  
        \end{align*}
        Using the fact that $\|e_j\|_2=1$ for all $j$ and $\|f\|_2=1$, we obtain
        \begin{align*}
            \left|\sum_{k_1 + \dots + k_n = \Delta - n} \lambda(k_1,\dots, k_n) \right|  & \leq \sum_{t=0}^{\Delta-n-1} \left| \sum_{k_1 + \dots + k_n =t} \lambda(k_1,\dots, k_n)\right|.
        \end{align*}
        This completes the proof for the proposition.

 \subsubsection{Proof of Proposition \ref{prop: equalitiy}}

            The KKT condition for the exact recovery that is the necessary and sufficient condition can be restated as 
        \begin{align}
         \exists \gamma_i \in \partial \| 0 \|_\circ, i \not \in \calK \text{ s.t. } \sum_{i \not \in \mathcal{K}} x_i \otimes \gamma_i =
        \sum_{i \in \mathcal{K}} x_i \otimes \partial \|\bar d_i \|_\circ.  \label{eq: kkt-fire}
        \end{align}
        For the problem \eqref{eq:hard-lasso2} with $\circ = 2$. the condition \eqref{eq: kkt-fire} becomes 
        \begin{align*}
         \exists \gamma_i \in \partial \| 0 \|_2, i \not \in \calK \text{ s.t. } \sum_{i \not \in \mathcal{K}} x_i \otimes \gamma_i =
        \sum_{i \in \mathcal{K}} x_i \otimes \partial \|\bar d_i \|_2 .
        \end{align*}
        Since $\frac{1}{\sqrt{n}} \partial \| 0 \|_1  =\frac{1}{\sqrt{n}}\mathbb{B}_\infty(1) \subseteq \mathbb{B}_2(1) = \partial \| 0 \|_2$, we can rewrite it as 
        \begin{align*}
         \exists \gamma_i \in \partial \| 0 \|_1, i \not \in \calK \text{ s.t. } \sum_{i \not \in \mathcal{K}} \frac{x_i}{\sqrt{n}}\otimes \gamma_i =
        \sum_{i \in \mathcal{K}} x_i \otimes \partial \|\bar d_i \|_2  .
        \end{align*}
        We can check the condition at each coordinate because the set $\mathbb{B}_\infty(1)$ is coordinate wise separable. Thus, the condition becomes that KKT condition holds for \eqref{eq:hard-lasso2} if there exist scalars $\gamma_i^l \in [-1,1], i \not \in \mathcal{K}, l = 1, \dots, n$ such that 
                \begin{align*}
                     \sum_{i \not \in \mathcal{K}} \gamma_i^l x_i/\sqrt{n}  =
        \sum_{i \in \mathcal{K}}  \partial \| \bar d_i \|_2^l  x_i, \quad \forall l=1,\dots, n ,
                \end{align*}
        where $\partial \| \bar d_i \|_\circ^l$ is the $l$-th element of the subgradient. Similar algebraic manipulation can be done for  \eqref{eq:hard-lasso1} except for the transforming subdifferential of the $\ell_2$ norm to subdifferential of the $\ell_1$ norm to obtain the second part of the result.

    \subsubsection{Proof of Lemma \ref{lem: farkas}}

            The condition "Given a matrix $\mathbf{F} \in \bR^{n \times m}$ and the vector $g \in \bR^n$, there exists a vector $w \in \bR^m$ with $\|w\|_\infty \le 1$ satisfying $\mathbf{F} w = g$." is equivalent to the feasibility of the linear programming (LP) below with objective function equal to $0$:
            \begin{align*}
                \max_{w \in \bR^m} \quad & 0 \\
                \text{ s.t. } ~& \mathbf{F} w = g, \\
                & \| w \|_\infty \le 1.
            \end{align*}
            Due to the strong duality, the dual problem of the LP above must have the optimum objective value equal to $0$. The dual problem can be formulated as 
            \begin{align*}
                \min_{y \in \bR^m,z \in \bR^n }  \quad & z^T g + \| y^T \|_1 \\
                \text{ s.t. } ~& z^T \mathbf{F}  + y^T = 0,
            \end{align*}
            or equivalently, 
            \begin{align*}
                \min_{z \in \bR^n }  \quad & f(z) := z^T g + \| z^T \mathbf{F}  \|_1 .
            \end{align*}
            Thus, for any $z \in \bR^n$, $f(z)$ must be nonnegative, i.e., $f(z) \ge 0$. Because $f(cz) = cf(z)$ for all $c>0$, the condition $f(z) \ge 0$ for all $z\in\bR^n$ is satisfied if $f(z) \ge 0$ for all $z \in \bR^n$ such that $\| z \|_2 = 1 $. This completes the proof.

    \subsubsection{Proof of Theorem \ref{thm: exact-l2-general-no-input}}

        Due to the system dynamics and given $x_0 = 0$, $x_i$ can be expressed as
        \[ x_i = \sum_{k \in \mathcal{K}} \bar A^{(i-k-1)_+}\bar d_k,\]
        where $A^{(i)_+}$ is defined as
        \begin{align*}
            A^{(i)_+} := \begin{cases} 0, \quad &\text{if } i< 0 \\ I, \quad  &\text{if } i= 0 \\ A^i, \quad &\text{if } i> 0
            \end{cases}.
        \end{align*}
        By Lemma \ref{lem: farkas}, given a coordinate $l\in\{1,\dots,n\}$, the optimality condition for the recovery of $\bar{A}$ is equivalent to 
        \begin{align}\label{eq:main-equation-1}
            f(z) := z^T g + \|z^T \mathbf{F} \|_1 \ge 0,\quad \forall z \in \mathbb{S}_2(1),
        \end{align}
        where the unit sphere $\mathbb{S}_2(1)$ is $\{ z \in \bR^n : \|z\|_2 = 1\}$, the matrix $\mathbf{F} \in \mathbb{R}^{n \times (T-|\mathcal{K}|)}$ has the columns 
        \[ \mathbf{F}^i :=  \sum_{k \in \mathcal{K}} \frac{\bar{A}^{(i-k-1)_+} \bar{d_k}}{\sqrt{n}}, \quad \forall i \not \in \calK, \]
        and the vector $g \in \bR^{n}$ is 
        \[ g := \sum_{i \in \mathcal{K}} \sum_{k \in \mathcal{K}} \bar{A}^{(i-k-1)_+} \bar{d_k} \cdot \bar f_i^l. \]
        We prove that condition \eqref{eq:main-equation-1} holds with high probability in two steps.

        \paragraph*{Step 1} We first prove that condition \eqref{eq:main-equation-1} holds with high probability for a fixed $z\in\mathbb{S}_2(1)$. 
        
        \paragraph{Step 1-1} We first analyze the term $\|z^T\mathbf{F}\|_1$, namely,
        \begin{align}\label{eq:step1-1}
            \mathbb{E} \|z^T\mathbf{F}\|_1 &= \frac{1}{\sqrt{n}}\sum_{i\notin\calK} \mathbb{E}\left| \sum_{k\in\calK} z^T \bar{A}^{(i-k-1)_+} \bar{d}_k \right|.
        \end{align}
        We construct the index set
        \[ \calI_1 := \{ i ~|~ i \notin\calK,~ i - 1 \in \calK \}. \]
        Let 
        \begin{align*} 
            S :=& \left\lceil \log_{\rho}\Theta\left[ \frac{c^{5}}{\log(|\calI_1|/\delta)} \right]\right\rceil\\
            =& \Theta\left[ \frac{\log\log( |\calI_1|/\delta ) + \log(1/c)}{\log(1/\rho)} \right], 
        \end{align*}
        where $\lceil x \rceil$ is the minimal integer that is not smaller than $x$ and $\delta\in(0,1)$ is the specified probability. We construct a subset of $\calI_1$ in the following way:
        \[ \calI := \{ i_1,\dots, i_I~|~ i_j\in\calI_1,~ i_j - i_{j-1} \geq S,~\forall j \}. \]
        It is straightforward to construct $\calI$ such that
        \[ I = |\calI| \geq \frac{1}{S} |\calI_1|. \]
        In addition, due to the probabilistic attack model, it holds with probability at least $1 - \exp[-\Theta[p(1-p)T]]$ that
        \[ |\calI_1| \geq \frac{p(1-p)T}2. \]
        Therefore, we have an estimate on the size of $\calI$:
        \begin{align}\label{eq:step1-2}
            \mathbb{P}\left(I \geq \frac{p(1-p)T}{2S} \right) \geq 1 - \exp[-\Theta[p(1-p)T]].
        \end{align}
        For each $j\in\{1,\dots,I\}$, we define
        \[ \calK_j := \{ k \in \calK ~|~ i_{j-1} < k < i_j \}, \]
        where we denote $i_0 := -1$. Moreover, we define
        \[ X_{j,\ell} := \sum_{k \in \calK_j} z^T\bar{A}^{i_\ell - k - 1}\bar{d}_k,\quad \forall j,\ell\in\{1,\dots,I\}\quad\mathrm{s.t.}~ j \leq \ell. \]
        Using equation \eqref{eq:step1-1}, we can calculate that
        \begin{align}\label{eq:step1-3}
            \|z^T\mathbf{F}\|_1 &\geq \frac{1}{\sqrt{n}}\sum_{\ell=1}^I\left|\sum_{j=1}^{\ell}X_{j,\ell}\right|\\
            \nonumber &\geq \frac{1}{\sqrt{n}}\sum_{j=1}^I\left(\left|X_{j,j}\right| - \sum_{\ell=j+1}^{I}\left|X_{j,\ell}\right|\right).
        \end{align}
        We utilize the following lemma to bound $|X_{j,\ell}|$.
        \begin{lemma}\label{lem:support-1}
        Suppose that a random variable $X$ is sub-Gaussian with parameter $\sigma_X$, where the mean and the variance of $X$ are $0$ and $\tilde{\sigma}_X^2$, respectively. Then, we have
        %
        \begin{align*}
            \mathbb{P}\left(|X| \geq \tilde{\sigma}_X\right) &\geq \frac{\tilde{\sigma}_X^4}{64\sigma_X^4}.
        \end{align*}
        \end{lemma}
        
        For all $j \in \{1,\dots,I\}$, the stealthy assumption (Assumption \ref{asp:stealth}) implies that the standard deviation and the sub-Gaussian parameter of $X_{j,\ell}$ are
        \begin{align*}
            \tilde{\sigma}_{j,\ell} &:= \sqrt{\frac1n\sum_{k \in \calK_j} \|z^T\bar{A}^{i_\ell - k - 1}\|_2^2\sigma_k^2},\\
            {\sigma}_{j,\ell} &:= \sqrt{\frac1n\sum_{k \in \calK_j} \|z^T\bar{A}^{i_\ell - k - 1}\|_2^2\sigma^2},
        \end{align*}
        respectively. 
        It follows from Lemma \ref{lem:support-1} that
        \begin{align*}
            \mathbb{P}\left(|X_{j,j}| \geq \tilde{\sigma}_{j,j}\right) &\geq \frac{\tilde{\sigma}_{j,j}^4}{64{\sigma}_{j,j}^4},
        \end{align*}
        which further leads to
        \begin{align}\label{eq:step1-4}
            \mathbb{P}\left(|X_{j,j}| \geq c{\sigma}_{j,j}\right) &\geq \frac{c^4}{64}.
        \end{align}
        %
        On the other hand, the sub-Gaussian parameter of $\sum_{\ell=j+1}^{I}\left|X_{j,\ell}\right|$ is at most
        \begin{align*}
            \sum_{\ell=j+1}^I \sigma_{j,\ell} \leq \sum_{\ell=j+1}^I \rho^{(\ell-j)S}\sigma_{j,j} \leq \frac{\rho^S}{1-\rho^S} \sigma_{j,j}.
        \end{align*}
        Therefore, it holds with probability at least $1 - \delta/(4I)$ that
        \begin{align}\label{eq:step1-6}
            - \sum_{\ell=j+1}^{I}\left|X_{j,\ell}\right| &\geq - \frac{\rho^S}{1-\rho^S} \sigma_{j,j} \cdot \sqrt{2\log(4I/\delta)}\\
            \nonumber&\geq - \frac{\rho^S}{1-\rho^S} \sigma_{j,j} \cdot \sqrt{2\log(4|\calI_1|/\delta)}\\
            \nonumber&\geq - \frac{c^4}{512} \cdot c\sigma_{j,j},
        \end{align}
        where the last step is by the choice of $S$. Using the bound in \eqref{eq:step1-2}, if we choose
        \begin{align*}
            T \geq \Theta\left(\frac{\log\log(1/\delta) + \log(1/c)}{p(1-p) c^4 \log(1/\rho)}\right),
        \end{align*}
        it holds with high probability that
        \[ \frac{c^4}{64} - \frac{\delta}{4I} \geq \frac{c^4}{128} . \]
        Note that we have dropped the $|\calI_1|$ term in the definition of $S$ since $\log\log(|\calI_1|)$ is bounded by $\log\log(T)$ and will not change the order of the above bound. Let $q_j$ be the $(1 - c^4 / 128)$-quantile of $\left|X_{j,j}\right| -\sum_{\ell=j+1}^{I}\left|X_{j,\ell}\right|$. 
        We define the indicator function
        \[ \mathbf{1}_j := \begin{cases}
            1,~ \text{if } \left|X_{j,j}\right| -\sum_{\ell=j+1}^{I}\left|X_{j,\ell}\right| \geq q_j,\\
            0,~ \text{otherwise},
        \end{cases}\quad \forall j \in \{1,\dots,I\}. \]
        Since the value of the Bernoulli random variable $\mathbf{1}_j$ only depends on attacks in $\calK_j$, which are disjoint from each other, the random variables 
        \[ \mathbf{1}_1 - c^4 / 128,~\dots,~\mathbf{1}_I - c^4 / 128 \]
        form a martingale sequence with respect to filtration $\calF_{i_1},\dots,\calF_{i_I}$. For all $j\in\{1,\dots,I\}$, we can calculate that
        \begin{align*}
            \mathbb{E}\left[ \exp\left( s \mathbf{1}_j \right) \right] \leq \exp\left[ \frac{c^4}{128}\left( e^s - 1 \right) \right],\quad \forall s\in\mathbb{R}.
        \end{align*}
        By the tower property of expectation, we have
        \begin{align*}
            \mathbb{E}\left[ \exp\left( s \sum_{j=1}^I\mathbf{1}_j \right) \right] \leq \exp\left[ \frac{c^4 I}{128}\left( e^s - 1 \right) \right],\quad \forall s\in\mathbb{R}.
        \end{align*}
        Therefore, by applying Chernoff's bound and choosing $s := -\log(2)$, it follows that
        \begin{align*}
            \mathbb{P}\left( \sum_{j=1}^I \mathbf{1}_j \leq \frac{c^4}{256} \cdot I \right) &\leq \exp\left[ - \frac{c^4 I}{256} \cdot s + \frac{c^4 I}{128}\left( e^s - 1 \right) \right]\\
            &\leq \exp\left[ -\frac{c^4 I}{256} \cdot \log\left(\frac12 \right) - \frac{c^4 I}{128} \cdot \frac12 \right]\\
            &= \exp\left[ -\Theta\left( \frac{c^4}{128} \cdot I \right) \right].
        \end{align*}
        %
        Equivalently, we know
        \begin{align}\label{eq:step1-7}
            \mathbb{P}\left( \sum_{j=1}^I \mathbf{1}_j \geq \frac{c^4}{256} \cdot I \right) \geq 1 - \exp\left[-\Theta\left( \frac{c^4}{128} \cdot I \right) \right].
        \end{align}
        Furthermore, since $i_j - 1 \in \calK_j$, we can estimate that
        \begin{align*}
            \sigma_{j,j} \geq \sqrt{\frac1n \|z\|_2^2 \sigma^2 } = \frac{1}{\sqrt{n}} \sigma.
        \end{align*}
        By the definition of $q_j$ and $\mathbf{1}_j$, when the event in inequality \eqref{eq:step1-7} happens, inequalities \eqref{eq:step1-4} and \eqref{eq:step1-6} imply that
        \begin{align*}
            \|z^T\mathbf{F}\|_1 &\geq \frac{1}{\sqrt{n}}\sum_{j=1}^I\left( \left|X_{j,j}\right| - \sum_{\ell=j+1}^{I}\left|X_{j,\ell}\right| \right)\\
            &\geq \frac{1}{\sqrt{n}}\sum_{j=1}^I\Bigg[\frac{c^4}{256} \cdot c\sigma_{j,j} - \frac{c^4}{512} \cdot c\sigma_{j,j}\Bigg] \geq \frac{c^5\sigma}{512n} \cdot I
        \end{align*}
        holds with probability at least $1-\delta/4$. Hence, we obtain
        \begin{align}\label{eq:step1-0}
            \mathbb{P}\left[ \|z^T\mathbf{F}\|_1 \geq \frac{c^5\sigma}{512n} \cdot I \right] \geq 1 - \exp\left[-\Theta\left( c^4 I \right) \right] - \frac{\delta}{4}.
        \end{align}

        \paragraph{Step 1-2} For the term $z^Tg$, we can establish an upper bound on
        \begin{align*}
            &\mathbb{E}\left[ \exp\left( \lambda \cdot z^Tg \right) \right]\\
            =& \mathbb{E}\left[\exp\left( \lambda\sum_{k \in \mathcal{K}} \sum_{i \in \mathcal{K}} z^T\bar{A}^{(i-k-1)_+} \bar{d_k} \cdot \bar f_i^l \right)\right].
        \end{align*}
        Define the filtration
        \[ \calF^f := \sigma\{\bar{f}_t, t\in\calK\}. \]
        By the stealth assumption, for each $k\in\calK$, conditional on $\calF_k$ and $\calF^f$, we have
        \[ \bar{\ell}_k \text{ is sub-Gaussian with parameter }\sigma. \]
        Let $T'$ be the second last time instance in $\calK$. We have
        \begin{align}\label{eq:step1-7-1}
            &\mathbb{E}\left[\exp\left( \lambda\sum_{i \in \mathcal{K}} \sum_{k \in \mathcal{K}} z^T\bar{A}^{(i-k-1)_+} \bar{d_k} \cdot \bar f_i^l \right)\right]\\
            \nonumber=& \mathbb{E}\Bigg[\exp\left( \lambda\sum_{k \in \mathcal{K},k < T'} \sum_{i \in \mathcal{K}} z^T \bar{A}^{(i-k-1)_+} \bar{d_k} \cdot \bar f_i^l \right) \\
            \nonumber&\hspace{-1em} \times\mathbb{E}\left[ \exp\left(\lambda \sum_{i \in \mathcal{K}} z^T \bar{A}^{(i-1-T')_+} \bar{d_T'}\cdot \bar f_{i}^l\right) \Bigg|\calF_{T'},\calF^f \right] \Bigg].
        \end{align}
        %
        %
        Using the decomposition in Assumption \ref{asp:stealth}, we have
        \begin{align*}
            &\mathbb{E}\left[ \exp\left(\lambda \sum_{i \in \mathcal{K}} z^T \bar{A}^{(i-1-T')_+} \bar{d_{T'}}\cdot \bar f_{i}^l\right) \Bigg|\calF_{T'},\calF^f \right]\\
            =& \mathbb{E}\left[ \exp\left(\lambda \sum_{i \in \mathcal{K}} z^T \bar{A}^{(i-1-T')_+} \bar{f}_{T'} \bar f_{i}^l \cdot \bar{\ell}_{T'} \right) \Bigg|\calF_{T'},\calF^f \right]\\
            \leq&\exp\left[\frac{\lambda^2\sigma^2}{2} \left(\sum_{i \in \mathcal{K}} z^T \bar{A}^{(i-1-T')_+} \bar{f}_{T'} \bar f_{i}^l\right)^2 \right].
        \end{align*}
        Substituting back into \eqref{eq:step1-7-1}, it follows that
        \begin{align*}
            &\mathbb{E}\left[\exp\left( \lambda\sum_{i \in \mathcal{K}} \sum_{k \in \mathcal{K}} z^T\bar{A}^{(i-k-1)_+} \bar{d_k} \cdot \bar f_i^l \right)\right]\\
            \nonumber\leq& \mathbb{E}\Bigg[\exp\left( \lambda\sum_{k \in \mathcal{K},k < T'} \sum_{i \in \mathcal{K}} z^T \bar{A}^{(i-k-1)_+} \bar{d_k} \cdot \bar f_i^l \right) \\
            \nonumber&\hspace{1em} \times\exp\left[\frac{\lambda^2\sigma^2}{2} \left(\sum_{i \in \mathcal{K}} z^T \bar{A}^{(i-1-T')_+} \bar{f}_{T'} \bar f_{i}^l\right)^2 \right] \Bigg].
        \end{align*}
        Continuing the process for all $k\in\calK$, we obtain
        \begin{align}\label{eq:step1-8}
            &\mathbb{E}\left[\exp\left( \lambda\sum_{i \in \mathcal{K}} \sum_{k \in \mathcal{K}} z^T\bar{A}^{(i-k-1)_+} \bar{d_k} \cdot \bar f_i^l \right)\right]\\
            \nonumber\leq&\mathbb{E}\left[\exp\left[\frac{\lambda^2\sigma^2}2 \sum_{k\in\calK} \left(\sum_{i \in \mathcal{K}} z^T \bar{A}^{(i-1-k)_+} \bar{f}_k \bar f_{i}^l\right)^2 \right]\right]\\
            \nonumber\leq&\mathbb{E}\left[\exp\left[\frac{\lambda^2\sigma^2}2 \sum_{k\in\calK} \left(\sum_{i \in \mathcal{K}} \left| z^T \bar{A}^{(i-1-k)_+} \bar{f}_k\right| \right)^2 \right]\right],
        \end{align}
        where the last inequality holds because $\bar{f}_i^l$ is bounded in $[-1,1]$. For each $i,k\in\calK$, the value of $\left(z^T \bar{A}^{(i-1-k)_+} \bar{f}_k\right)^2$ concentrates around its expectation $ \|z^T \bar{A}^{(i-1-k)_+}\|_2^2 / n$. Therefore, inequality \eqref{eq:step1-8} leads to
        \begin{align}\label{eq:step1-8-1}
            &\mathbb{E}\left[\exp\left( \lambda\sum_{i \in \mathcal{K}} \sum_{k \in \mathcal{K}} z^T\bar{A}^{(i-k-1)_+} \bar{d_k} \cdot \bar f_i^l \right)\right]\\
            \nonumber\leq&\exp\left[\Theta\left[\frac{\lambda^2\sigma^2}{2n} \sum_{k\in\calK} \left(\sum_{i \in \mathcal{K}} \left\| z^T \bar{A}^{(i-1-k)_+}\right\|_2 \right)^2 \right]\right]\\
            \nonumber\leq&\exp\left[\Theta\left[\frac{\lambda^2\sigma^2}{2n} \sum_{k\in\calK} \left(\sum_{i \in \mathcal{K}} \rho^{(i-k-1)_+} \right)^2 \right]\right].
        \end{align}
        Suppose the elements in $\calK$ are
        \[ j_1 < j_2 < \dots < j_{|\calK|}. \]
        Define
        \[ \Delta_k := j_k - j_{k-1} - 1,\quad \forall k\in\{2,\dots,|\calK|\}. \]
        We can calculate that 
        \[ \sum_{i \in \mathcal{K}} \rho^{(i-1-j_k)_+} \leq \frac{\rho^{\Delta_{k}}}{1 - \rho}. \]
        Since $\rho^{\Delta_k}\in[0,1]$ are bounded random variables, they are sub-Gaussian and concentrate around the mean with high probability. The expectation of $\rho^{2\Delta_k}$ is 
        \[ \sum_{\Delta=0}^\infty p (1-p)^\Delta \rho^{2\Delta} = \frac{p}{1 - (1-p)\rho^2}. \]
        Therefore, with probability at least $1 - \exp[-\Theta(pT)]$, we have
        \begin{align*} 
            \sum_{k=2}^{|\calK|} \rho^{2\Delta_k} &\lesssim \frac{|\calK| p}{1 - (1-p)\rho^2} \leq \frac{|\calK| p}{1 - \rho}.
        \end{align*}
        Hence, inequality \eqref{eq:step1-8-1} implies that with the same probability, $z^Tg$ is sub-Gaussian with parameter
        \[ \Theta\left[\sqrt{\frac{\sigma^2}{n} \sum_{k=2}^{|\calK|}\frac{\rho^{2\Delta_{k}}}{(1 - \rho)^2}}\right] \leq \Theta\left[\sqrt{\frac{|\calK|p \sigma^2}{n(1 - \rho)^3}}\right]. \]
        Therefore, Hoeffding's inequality leads to
        \begin{align}\label{eq:step1-9}
            \mathbb{P}\left[ z^Tg \leq - \Theta\left( \sqrt{\frac{|\calK|p\sigma^2}{n(1-\rho)^3} \log\left(\frac{4}{\delta}\right)} \right) \right] \leq \frac{\delta}{4}.
        \end{align}

        By combining inequalities \eqref{eq:step1-0} and \eqref{eq:step1-9}, it holds with probability at least
        \begin{align*}
            1 - \exp\left[-\Theta(c^4I)\right] - \frac{\delta}2
        \end{align*}
        that 
        \begin{align*}
            f(z) \geq \Theta\left[ \frac{c^5\sigma I}{n} - \sqrt{\frac{|\calK|p\sigma^2}{n(1-\rho)^3} \log\left(\frac{1}{\delta}\right)} \right].
        \end{align*}
        Similar to the bound in \eqref{eq:step1-2}, it holds with probability at least $1 - \exp[-\Theta(pT)]$ that
        \[ |\calK| \leq 2pT. \]
        As a result, if we choose
        %
                \begin{align}\label{eq:step1-10}
            T \geq \Theta\Bigg[ \max\Bigg\{ &\frac{\log\log({1}/{\delta}) + \log(1/c)}{c^4 p(1-p) \log(1/\rho)} \log\left(\frac{1}{\delta}\right),\\
            \nonumber&\frac{1}{p(1-p)} \log\left(\frac{1}{\delta}\right),\\
            \nonumber&\frac{n\log(1/c)^2}{c^{10} (1-p)^2 (1-\rho)^3\log^2(1/\rho)} \log\left(\frac{1}{\delta}\right) \Bigg\} \Bigg]\\
            \nonumber&\hspace{-5em}= \Theta\Bigg[ nR \log\left(\frac{1}{\delta}\right) \Bigg],
        \end{align}
        where
        %
        %
        \begin{align*}
            R := \max \Bigg\{ & \frac{\log(1/c)}{c^4 p(1-p) \log(1/\rho)} \log\left(\frac{1}{\delta}\right), \\ 
            & \frac{\log^2(1/c)}{c^{10} (1-p)^2 (1-\rho)^3\log^2(1/\rho)}, \frac{1}{np(1-p)} \Bigg\},
        \end{align*}
        we have
        \begin{align}\label{eq:step1-11}
            \mathbb{P}\left[f(z) \geq \Theta\left(\frac{c^5\sigma I}{n}\right) \right] \geq 1 - \delta.
        \end{align}

        \paragraph*{Step 2} In the second step, we apply discretization techniques to prove that condition \eqref{eq:main-equation-1} holds for all $z\in\mathbb{S}_2(1)$ with high probability. Suppose that $\epsilon > 0$ is a small constant. We construct an $\epsilon$-cover of the unit sphere $\mathbb{S}_2(1)$, denoted as
        \[ \{z^1,\dots,z^N\}, \]
        Namely, for all $z \in \mathbb{S}_2(1)$, we can find $r \in \{1,2, \dots, N\}$ such that $\| z - z^r \|_2 \le \epsilon$. The number of points $N$ can be bounded by
        \begin{align*}
            \log(N) \leq \log [\calN (\epsilon, \mathbb{S}_2(1), \|\cdot\|_2)] & \le n\log \left(1 + \frac{2}{\epsilon}\right).
        \end{align*}
        Define $a$ to be the lower bound of $f(z)$ in inequality \eqref{eq:step1-11}. Then, we have
        \[ a = \Theta\left(\frac{c^5\sigma I}{n}\right). \]
        Our goal is to prove that
        \[ f(z) - f(z') \geq -{a},\quad \forall z,z' \in \mathbb{S}_2(1)~\mathrm{s.t.}~\|z-z'\|_2 \leq \epsilon \]
        holds with high probability. Notice that
        \begin{align*} 
            f(z) - f(z') &= (z-z')^Tg + ( \|z^T\mathbf{F}\|_1 - \|(z')^T\mathbf{F}\|_1 )\\
            &\geq (z-z')^Tg - \|(z-z')^T\mathbf{F}\|_1\\
            &\geq -\|z - z'\|_2\|g\|_2 - \|z - z'\|_2 \sum_{i\notin\calK} \|\mathbf{F}^i\|_2\\
            &\geq -\epsilon\Bigg( \left\|\sum_{i\in\calK} \sum_{k\in\calK} \bar{A}^{(i-k-1)_+}\bar{d}_k\right\|_2\\
            &\hspace{5em}+ \frac{1}{\sqrt{n}}\sum_{i\notin\calK} \left\|\sum_{k\in\calK} \bar{A}^{(i-k-1)_+}\bar{d}_k\right\|_2 \Bigg)\\
            &\geq -\epsilon\sum_{k\in\calK}\sum_{i>k} \rho^{(i-k-1)}|\bar{\ell}_k|.
        \end{align*}
        Using the property of exponential sequences, we have
        \[ \sum_{k\in\calK}\sum_{i>k} \rho^{(i-k-1)}|\bar{\ell}_k| \leq \frac{1}{1-\rho} \sum_{k\in\calK} |\bar{\ell}_k|. \]
        Using a similar proof, we can show that $\sum_{k\in\calK} |\bar{\ell}_k|$ is sub-Gaussian with parameter $|\calK|\sigma$. Therefore, Hoeffding's inequality implies that
        \begin{align*}
            \mathbb{P}\left( \frac{1}{1-\rho} \sum_{k\in\calK} |\bar{\ell}_k| > \frac{a}{\epsilon} \right) \leq 2\exp\left[ -\frac{(1-\rho)^2a^2}{2\epsilon^2|\calK|^2\sigma^2} \right].
        \end{align*}
        Letting
        \[ \epsilon := \frac{(1 - \rho)a}{|\calK|\sigma\sqrt{2\log(4/\delta)}}, \]
        it holds that 
        \begin{align*} 
            &\mathbb{P}\left[ f(z) - f(z') \geq -{a},\quad \forall z,z'\in\mathbb{S}_2(1)~\mathrm{s.t.}~\|z-z'\|_2\leq \epsilon \right]\\
            \geq& \mathbb{P}\left( \frac{1}{1-\rho} \sum_{k\in\calK} |\bar{\ell}_k| \leq \frac{a}{\epsilon} \right) \geq 1 - \frac{\delta}{2}.
        \end{align*}
        Now, after we replace $\delta$ in \eqref{eq:step1-10} with $\delta/(2N)$, it holds with probability at least $1-\delta/2$ that
        \begin{align*}
            f(z^r) \geq {a},\quad \forall r\in\{1,\dots,N\}.
        \end{align*}
        After combining the above two inequalities, we apply the union bound to obtain
        \[ \mathbb{P}\left[ f(z) \geq 0,\quad \forall z \in \mathbb{S}_2(1) \right] \geq 1 - \delta. \]
        The corresponding sample complexity is
        \begin{align*}
            T \geq \Theta\Bigg[ nR \log\left(\frac{2N}{\delta}\right) \Bigg].
        \end{align*}
        Since it holds with probability $1-\exp[-\Theta[p(1-p)T]]$ that
        \[ |\calI_1| = \Theta[p(1-p)T],\quad |\calK| = \Theta(pT), \]
        we get the estimate
        \begin{align*}
            \log(N) &\leq n \log\left(1 + \frac{2}{\epsilon} \right)\\
            &= n \log\left[ 1 + \Theta\left(\frac{n\sqrt{\log(1/\delta)}\log(1/c)}{(1-p)c^5(1-\rho)\log(1/\rho)}\right) \right]\\
            &=\Theta\left[n \log\left(nR\right) \right].
        \end{align*}
        By omitting the constants in the expression, the final sample complexity can be written as
        \begin{align*}
            T \geq \Theta\Bigg[ nR \left[n\log\left({nR}\right) + \log\left(\frac{1}{\delta}\right)\right]\Bigg]. 
        \end{align*}
        Finally, we replace $\delta$ with $\delta/n$ and apply the union bound to all coordinates $\ell\in\{1,\dots,n\}$. The sample complexity remains on the same order as the above expression.

    \subsubsection{Proof of Theorem \ref{thm: general-l2}}

    Due to the system dynamics and given $x_0 = 0$, $x_i$ can be expressed as
    \[ x_i = \sum_{k \notin\calK } \bar A^{(i-k-1)_+}\bar B u_k + \sum_{k \in \mathcal{K}} \bar A^{(i-k-1)_+}(\bar{B}u_k + \bar d_k). \]
    From Proposition \ref{prop: general}, we want to show that there exist scalars $\gamma_i^l, \mu_i^l \in [-1,1]$ for all $ i \not \in \mathcal{K}, l \in \{1, \dots, n\}$ such that 
    \begin{align}
             \sum_{i \not \in \mathcal{K}} \gamma_i^l x_i/\sqrt{n}  =
    \sum_{i \in \mathcal{K}}  \partial \| \bar d_i \|_2^l  x_i, \quad \forall l=1,\dots, n, \label{eq: state}          
            \end{align}
    and
            \begin{align}
                 \sum_{i \not \in \mathcal{K}} \mu_i^l u_i/\sqrt{n}  =
    \sum_{i \in \mathcal{K}}  \partial \| \bar d_i \|_2^l  u_i, \quad \forall l=1,\dots, n.\label{eq: input} 
            \end{align}
    We finish the proof in two steps.
    
    \paragraph*{Step 1} We first analyze condition \eqref{eq: state} with a given coordinate $l\in\{1,\dots,n\}$. From Lemma \ref{lem: farkas}, condition \eqref{eq: state} is equivalent to 
    \[ f(z) := z^T g + \|z^T \mathbf{F} \|_1 \ge 0,\quad \forall z \in \mathbb{S}_2(1), \]
    where the matrix $\mathbf{F} \in \mathbb{R}^{n \times (T-|\mathcal{K}|)}$ has the columns 
    \[ \mathbf{F}^i := \sum_{k \notin \calK} \frac{\bar A^{(i-k-1)_+} \bar B u_k}{\sqrt{n}}  +  \sum_{k \in \mathcal{K}} \frac{\bar A^{(i-k-1)_+} (\bar{B}u_k + \bar d_k)}{\sqrt{n}}, ~ \forall i \not \in \calK, \]
    and the vector $g \in \bR^{n}$ is 
    \[ g := \sum_{i \in \mathcal{K}} \left[ \sum_{k \notin \calK} \bar A^{(i-k-1)_+}  \bar B u_k  + \sum_{k \in \mathcal{K}} \bar A^{(i-k-1)_+} (\bar{B}u_k + \bar d_k) \right] \bar f_i^l. \]
    Similar to the proof of Theorem \ref{thm: exact-l2-general-no-input}, we first prove that $f(z)\geq a$ holds with high probability for a fixed $z\in\mathbb{S}_2(1)$ and some positive constant $a$. For each $k\notin\calK$, the standard deviation and sub-Gaussian parameter of $z^T\bar A^{(i-k-1)_+} \bar B u_k$ are both
    \[ \frac{1}{\sqrt{m}}\|z^T\bar A^{(i-k-1)_+} \bar B\|_2 \xi. \]
    For each $k\in\calK$, the standard deviation and sub-Gaussian parameter of $z^T\bar A^{(i-k-1)_+} (\bar B u_k + \bar d_k)$ are, respectively,
    \begin{align*}
        &\sqrt{ \frac1m\|z^T\bar A^{(i-k-1)_+} \bar B\|_2^2 \xi^2 + \frac1n \|z^T\bar A^{(i-k-1)_+}\|_2^2 \sigma_k^2 },\\
        &\sqrt{ \frac1m\|z^T\bar A^{(i-k-1)_+} \bar B\|_2^2 \xi^2 + \frac1n \|z^T\bar A^{(i-k-1)_+}\|_2^2 \sigma^2 }. 
    \end{align*}
    Note that we have utilized the independence between $u_k$ and $\bar{d}_k$ in the above calculation. Let 
    \[ S := \left\lceil \log_\rho \Theta\left[ \frac{\sqrt{\frac{1}{m}\eta_B^2\xi^2 + \frac{p}{n}\sigma^2} \cdot c^5 }{\sqrt{\frac{1}{m}\rho_B^2\xi^2 + \frac{p}{n}\sigma^2} \cdot \sqrt{\log(1/\delta)}} \right]\right\rceil, \]
    where $\rho_B$ is the maximal singular value of $\bar{B}$ and $\eta_B$ is the minimal singular value of the matrix
    \begin{align*}
        \frac{1}{(1-\rho)^2} \begin{bmatrix}
            \bar{B} & \bar{A}\bar{B} &\cdots & \bar{A}^{n-1}\bar{B}
        \end{bmatrix}.
    \end{align*}
    By the controllability assumption, the above matrix is rank-$n$ and thus, the parameter $\eta_B$ is strictly positive. We define $i_0 := -1$ and construct the index set 
    \begin{align*}
        \calI := \{ i_1,\dots,i_I ~|~ i_j\notin \calK,~ i_{j} - i_{j-1}\geq S,\quad\forall j \}.
    \end{align*}
    It is straightforward to construct $\calI$ such that $I = |\calI|$ is on the order of
    \[ \min\left\{ (1-p)T, \frac{T}{S} \right\}. \]
    %
    For each $j\in\{1,\dots,I\}$, we define
    \[ \calK_j := \{ k \in \calK ~|~ i_{j-1} \leq k < i_j \},~ \calK_j^c := \{ k \notin \calK ~|~ i_{j-1} \leq k < i_j \}. \]
    Moreover, we define
    \begin{align*} 
        X_{j,\ell} := &\sum_{k\in\calK_j} z^T\bar{A}^{i_\ell - k - 1}\left(\bar{B}u_k + \bar{d}_k\right) + \sum_{k\in\calK_j^c} z^T\bar{A}^{i_\ell - k - 1}\bar{B}u_k,\\
        &\hspace{10em} \forall j,\ell\in\{1,\dots,I\}\quad \mathrm{s.t.}~j\leq \ell.
    \end{align*}
    For all $j \in \{1,\dots,I\}$, the stealthy assumption (Assumption \ref{asp:stealth}) implies that the standard deviation and the sub-Gaussian parameter of $X_{j,\ell}$ is
    \begin{align*}
        &\quad\tilde{\sigma}_{j,\ell}:=\\
        & \sqrt{\frac1m\sum_{k\in\calK_j\cup\calK_j^c} \|z^T\bar{A}^{i_\ell - k - 1}\bar{B}\|_2^2\xi^2 + \frac1n\sum_{k\in\calK_j} \|z^T\bar{A}^{i_\ell - k - 1}\|_2^2\sigma_k^2},\\
        &\quad{\sigma}_{j,\ell}:=\\
        & \sqrt{\frac1m\sum_{k\in\calK_j\cup\calK_j^c} \|z^T\bar{A}^{i_\ell - k - 1}\bar{B}\|_2^2\xi^2 + \frac1n\sum_{k\in\calK_j} \|z^T\bar{A}^{i_\ell - k - 1}\|_2^2\sigma^2},
    \end{align*}
    respectively. Define
    \[ c_{j,\ell} := \frac{\tilde{\sigma}_{j,\ell}}{\sigma_{j,\ell}},\quad \forall j,\ell\in\{1,\dots,I\}\quad \mathrm{s.t.}~j\leq \ell. \]
    Similar to the proof of Theorem \ref{thm: exact-l2-general-no-input}, we have the bound
    \begin{align*}
    \|z^T\mathbf{F}\|_1 \geq \frac{1}{\sqrt{n}} \sum_{j=1}^I\left( |X_{j,j}| - \sum_{\ell = j+1}^I |X_{j,\ell}| \right).
    \end{align*}
    By Lemma \ref{lem:support-1}, we have
    \begin{align}\label{eq:input1-1}
        \mathbb{P}\left( |X_{j,j}| \geq c_{j,j}{\sigma}_{j,j} \right) &\geq \frac{c_{j,j}^4}{64}.
    \end{align}
    For all vector $y\in\mathbb{R}^n$, the controllability assumption leads to
    \begin{align}\label{eq:input1-1-1}
        &\sum_{k=0}^{n-1}\|y^T \bar{A}^k \bar{B}\|_2^2 \geq \frac{\eta_B^2}{(1-\rho)^2} \cdot \|y\|_2^2\\
        \nonumber\geq& \frac{\eta_B^2}{(1-\rho)^2} \cdot (1-\rho)^2 \sum_{k=0}^n \rho^{2k} \|y\|_2^2 \geq \eta_B^2 \sum_{k=0}^n \|y^T \bar{A}^k\|_2^2.
    \end{align}
    Therefore, we can divide the set $\calK_j\cup\calK_j^c$ into segments with $n$ consecutive time instances and apply inequality \eqref{eq:input1-1-1} to each segment. When $T$ is large enough such that $I \geq \Theta(n)$, we obtain the estimation
    \begin{align*}
        \sum_{k\in\calK_j\cup\calK_j^c} \|z^T\bar{A}^{i_\ell - k - 1}\bar{B}\|_2^2 &\gtrsim \sum_{k=i_{j-1}}^{i_j - 1} \|z^T\bar{A}^{i_\ell - k - 1}\|_2^2\eta_B^2.
    \end{align*}
    Applying concentration inequalities to set $\calK_j$, the distribution of its elements will surround their expected values. Therefore, for the simplicity of presentation, we use the following approximation:
    \[ {\sigma}_{j,j}^2 \gtrsim \frac1m \eta_B^2 \xi^2 + \frac{p}n \sigma^2 := \bar\sigma^2. \]
    In addition, the parameter $c_{j,j}$ can be bounded by
    \begin{align*}
        &\quad c_{j,j}^2\\
        &\geq \frac{\frac1m\sum_{k\in\calK_j\cup\calK_j^c} \|z^T\bar{A}^{i_\ell - k - 1}\bar{B}\|_2^2\xi^2}{\frac1m\sum_{k\in\calK_j\cup\calK_j^c} \|z^T\bar{A}^{i_\ell - k - 1}\bar{B}\|_2^2\xi^2 + \frac1n\sum_{k\in\calK_j} \|z^T\bar{A}^{i_\ell - k - 1}\|_2^2\sigma^2}\\
        &= \Bigg[1 + \left(\frac1m\sum_{k\in\calK_j\cup\calK_j^c} \|z^T\bar{A}^{i_\ell - k - 1}\bar{B}\|_2^2\xi^2\right)^{-1}\\
        &\hspace{13em}\times\frac1n\sum_{k\in\calK_j} \|z^T\bar{A}^{i_\ell - k - 1}\|_2^2\sigma^2\Bigg]^{-1}.
    \end{align*}

    For the numerator, we can estimate that
    \begin{align*}
        \sum_{k\in\calK_j\cup\calK_j^c} \|z^T\bar{A}^{i_\ell - k - 1}\bar{B}\|_2^2 &\gtrsim \sum_{k=i_{j-1}}^{i_j - 1} \|z^T\bar{A}^{i_\ell - k - 1}\|_2^2\eta_B^2.
    \end{align*}
    On the other hand, since
    \begin{align*}
        \sum_{k\in\calK_j\cup\calK_j^c} \|z^T\bar{A}^{i_\ell - k - 1}\bar{B}\|_2^2\xi^2\leq &\sum_{k\in\calK_j\cup\calK_j^c} \|z^T\bar{A}^{i_\ell - k - 1}\|_2^2 \rho_B^2\xi^2,\\
        \sum_{k\in\calK_j} \|z^T\bar{A}^{i_\ell - k - 1}\|_2^2\sigma^2 \lesssim & p\sum_{k\in\calK_j\cup\calK_j^c} \|z^T\bar{A}^{i_\ell - k - 1}\|_2^2\sigma^2,
    \end{align*}
    we get
    \[ c_{j,j}^2 \gtrsim \frac{\frac1m \eta_B^2 \xi^2}{\frac1m \rho_B^2 \xi^2 + \frac{p}n \sigma^2} := c. \]
    Therefore, inequality \eqref{eq:input1-1} implies
    \begin{align}\label{eq:input1-2}
        \mathbb{P}(|X_{j,j}| \geq c\bar{\sigma}) \geq \frac{c^4}{64}.
    \end{align}
    Since the sub-Gaussian parameter of $\sum_{\ell = j+1}^I |X_{j,\ell}|$ is $\sum_{\ell = j+1}^I \sigma_{j,\ell}$, Hoeffding's inequality implies that

    \begin{align}\label{eq:input1-3}
        \mathbb{P}\left( \sum_{\ell = j+1}^I |X_{j,\ell}| \leq \sum_{\ell = j+1}^I \sigma_{j,\ell} \cdot \sqrt{2\log\left(\frac{4I}{\delta}\right)} \right) &\geq 1 - \frac{\delta}{4I}.
    \end{align}
    We can bound the sub-Gaussian parameter by
    \begin{align*}
        &\quad\sum_{\ell = j+1}^I \sigma_{j,\ell}\\
        &\leq \sum_{\ell = j+1}^I\sqrt{\frac1m\sum_{k\in\calK_j\cup\calK_j^c} \rho^{2(i_\ell - k - 1)}\rho_B^2\xi^2 + \frac1n\sum_{k\in\calK_j} \rho^{2(i_\ell - k - 1)}\sigma^2}\\
        &\leq \frac{\rho^S}{1-\rho^S}\sqrt{\frac1m\sum_{k\in\calK_j\cup\calK_j^c} \rho^{2(i_j - k - 1)}\rho_B^2\xi^2 + \frac1n\sum_{k\in\calK_j} \rho^{2(i_j - k - 1)}\sigma^2}\\
        &\leq \frac{\rho^S}{1-\rho^S}\sqrt{\frac{1}{m(1-\rho)}\rho_B^2\xi^2 + \frac1n\sum_{k\in\calK_j} \rho^{2(i_j - k - 1)}\sigma^2}.
    \end{align*}
    In the same way, we have the following bound with high probability:
    \begin{align*}
        \sum_{k\in\calK_j} \rho^{2(i_j - k - 1)} \lesssim p \sum_{k\in\calK_j\cup\calK_j^c} \rho^{2(i_j - k - 1)} \leq \frac{p}{1-\rho},
    \end{align*}
    which holds with high probability when $T$ is large. Therefore, we have the bound
    \begin{align*}
        \sum_{\ell = j+1}^I \sigma_{j,\ell} &\lesssim \frac{\rho^S}{1-\rho^S}\sqrt{\frac{1}{m(1-\rho)}\rho_B^2\xi^2 + \frac{p}{n(1-\rho)}\sigma^2}\\
        &:= \frac{\rho^S}{1-\rho^S} \tilde{\sigma}.
    \end{align*}
    By the choice of $S$, we get
    \begin{align*}
        \sum_{\ell = j+1}^I \sigma_{j,\ell} \lesssim \frac{c^4}{256} \cdot c\bar{\sigma} \cdot \left(\sqrt{2\log\left(\frac{4I}{\delta}\right)}\right)^{-1},
    \end{align*}
    Therefore, inequality \eqref{eq:input1-3} leads to
    \begin{align}
        \mathbb{P}\left( \sum_{\ell = j+1}^I |X_{j,\ell}| \leq \frac{c^4}{256} \cdot c\bar{\sigma} \right) &\geq 1 - \frac{\delta}{4I}.
    \end{align}
    Choosing
    \[ T \geq \Theta\left( \frac{\log\log(1/\delta)}{c^4\min\{1-p, 1/S\}} \right), \]
    we have
    \[ \frac{c^4}{64} - \frac{\delta}{4I} \geq \frac{c^4}{128}. \]
    By the same construction of the martingale sequence and the application of Azuma-Hoeffding's inequality, inequalities \eqref{eq:input1-1} and \eqref{eq:input1-3} imply that
    \begin{align}\label{eq:input1-4}
        \|z^T\mathbf{F}\|_1 &\geq \frac{1}{\sqrt{n}}\sum_{j=1}^I\left( |X_{j,j}| - \sum_{\ell = j + 1}^I|X_{j,\ell}| \right)\\
        \nonumber&\geq \frac{1}{\sqrt{n}}\left( \frac{c^4I}{256} \cdot c\bar{\sigma} - \frac{c^4I}{512} c\bar{\sigma} \right) = \frac{c^5 \bar{\sigma}}{512\sqrt{n}} \cdot I
    \end{align}
    holds with probability at least
    \[ 1- \exp[-\Theta(c^4I)]-\delta/4. \]

    On the other hand, for the term $z^Tg$, we can bound its sub-Gaussian parameter by
    \begin{align*}
        \sqrt{\frac{|\calK|\rho_B^2\xi^2}{m(1-\rho)^3} + \frac{|\calK| p\sigma^2}{n(1-\rho)^3}} = \sqrt{\frac{|\calK|}{(1-\rho)^2}} \tilde{\sigma} .
    \end{align*}
    Then, Hoeffding's inequality leads to
    \begin{align}\label{eq:input1-5}
    \mathbb{P}\left[ z^Tg \leq - \Theta \left(\sqrt{\frac{|\calK|\tilde{\sigma}^2}{(1-\rho)^2}\log\left(\frac{4}{\delta}\right)} \right) \right] \leq \frac{\delta}{4}.
    \end{align}

    Combining inequalities \eqref{eq:input1-4} and \eqref{eq:input1-5}, it holds with probability at least
    \[ 1 - \exp[-\Theta(c^4 I)] - \frac{\delta}{2} \]
    that
    \[ f(z) \geq \Theta\left[ \frac{c^5\bar{\sigma} I}{\sqrt{n}} - \sqrt{\frac{\tilde{\sigma}^2|\calK|}{(1-\rho)^2}\log\left(\frac{1}{\delta}\right)} \right]. \]
    Similar to the bound in \eqref{eq:step1-2}, it holds with probability at least $1 - \exp[-\Theta(pT)]$ that
    \[ |\calK| \leq 2pT. \]
    As a result, if we choose
    \begin{align*}
        T \geq \Theta\Bigg[ \max\Bigg\{ &\frac{1}{c^4 \min\{1-p, 1/S\}} \log\left(\frac{1}{\delta}\right),\\
        \nonumber&\frac{1}{p} \log\left(\frac{1}{\delta}\right),\\
        \nonumber&\frac{np \kappa^2}{c^{10}(1-\rho)^2 \min\{(1 - p)^2, 1/S^2\} } \log\left(\frac{1}{\delta}\right) \Bigg\} \Bigg]\\
        \nonumber&\hspace{-5em}= \Theta\Bigg[ nR_1 \log\left(\frac{1}{\delta}\right) \Bigg],
    \end{align*}
    where $\kappa := \tilde{\sigma}/\bar{\sigma} \geq (1-\rho)^{-1}$ and
    \begin{align*}
        R_1 := \max \Bigg\{ & \frac{1}{c^4  (1-p)},  \frac{\log(\kappa/c)}{c^4  \log(1/\rho)}, \\
         & \frac{p \kappa^2}{c^{10} (1 - p)^2 (1-\rho)^2},\frac{p \kappa^2\log^2(\kappa/c)}{c^{10}(1-\rho)^2 \log^2(1/\rho) },\frac{1}{np} \Bigg\}
    \end{align*}
    we have
    \begin{align}\label{eq:input1-6}
        \mathbb{P}\left[f(z) \geq \Theta\left(\frac{c^5\bar{\sigma} I}{\sqrt{n}}\right) \right] \geq 1 - \delta.
    \end{align}

    Next, we apply the discretization techniques and estimate the size of $\epsilon$-net, which we denote as $N$. Similar to the proof of Theorem \ref{thm: exact-l2-general-no-input}, it is sufficient to choose
    \[ \log(N) \le n \log\left( 1 + \frac{2}{\epsilon} \right), \]
    and
    \[ \epsilon := \Theta\left(\frac{a}{\|g\|_2 + \sum_{i\notin\calK}\|\mathbf{F}^i\|_2} \right). \]
    where $a > 0$ is the lower bound of $f(z)$ in \eqref{eq:input1-6}. We can estimate that
    \begin{align*}
        \|g\|_2 &\leq \sum_{i\in\calK}\left\| \sum_{k \notin \calK} \bar A^{(i-k-1)_+}  \bar B u_k  + \sum_{k \in \mathcal{K}} \bar A^{(i-k-1)_+} (\bar{B}u_k + \bar d_k) \right\|_2\\
        &\leq \frac{\rho_B}{1 - \rho} \sum_{k=0}^{T-1}\|u_k\|_2 + \frac{1}{1 - \rho} \sum_{j\in\calK} \|\bar{d}_k\|_2.
    \end{align*}
    Therefore, the sub-Gaussian parameter of $\|g\|_2$ is bounded by
    \begin{align*}
        \frac{1}{1 - \rho}\sqrt{\rho_B^2 T\xi^2 +  |\calK|\sigma^2} \lesssim \frac{1}{1 - \rho}\sqrt{\rho_B^2 T\xi^2 +  pT\sigma^2} := \sigma'\sqrt{T}.
    \end{align*}
    Similarly, the sub-Gaussian parameter of $\sum_{i\notin\calK}\|\mathbf{F}^i\|_2$ is bounded by
    \begin{align*}
        \frac{1}{(1 - \rho)\sqrt{n}}\sqrt{\rho_B^2 T\xi^2 +  pT\sigma^2} = \frac{\sigma'\sqrt{T}}{\sqrt{n}}
    \end{align*}
    with high probability. Hoeffding's inequality implies
    \begin{align*}
        \|g\|_2 + \sum_{i\notin\calK}\|\mathbf{F}^i\|_2 \leq \Theta \left[ \sigma' \sqrt{T} \sqrt{\log\left(\frac{1}{\delta}\right)} \right]
    \end{align*}
    with probability at least $1 - \delta/2$. With the same probability, we have
    \begin{align*}
        \log\left(1 + \frac{2}{\epsilon}\right) &\leq \log\left[1 + \Theta\left( \frac{\sqrt{n}\sigma'\sqrt{\log(1/\delta)}}{c^5 \bar{\sigma}\min\{ 1-p, 1/S \} \sqrt{T} } \right) \right]\\
        &\leq \Theta[\log(nR_1)],
    \end{align*}
    which further leads to
    \[ \log(N) \lesssim \Theta[n\log(nR_1)]. \]
    Replacing $\delta$ with $\delta / N$ in \eqref{eq:input1-6}, the final sample complexity bound is
    \begin{align*}
        T \geq \Theta\left[ n R_1 \left[ n\log(nR_1) + \log\left(\frac{1}{\delta}\right) \right] \right].
    \end{align*}

    \paragraph*{Step 2} In the second step, we consider condition \eqref{eq: input}. From Lemma \ref{lem: farkas}, given a coordinate $l\in\{1,\dots,n\}$, \eqref{eq: input} is equivalent to 
    \[ f(z) := z^T g + \|z^T \mathbf{F} \|_1 \ge 0,\quad \forall z \in \mathbb{S}_2(1), \]
    where the matrix $\mathbf{F} \in \mathbb{R}^{m \times (T-|\mathcal{K}|)}$ has the columns 
    \[ \mathbf{F}^i :=  \frac{u_i}{\sqrt{n}} , \quad \forall i \not \in \calK, \]
    and the vector $g \in \bR^{m}$ is 
    \[ g := \sum_{i \in \mathcal{K}} u_i \bar{f}_i^l. \]
    For a given $z\in\mathbb{S}_2(1)$, we have
    \begin{align*} 
        \mathbb{E}[f(z)] &= \mathbb{E}\|z^T\mathbf{F}\|_1= \sum_{i\notin \calK}\mathbb{E}|z^T\mathbf{F}^i|\\
        &= \Theta\left(\frac{(T - |\calK|)\xi}{\sqrt{mn}} \right) \gtrsim \Theta\left(\frac{(1-p)T \xi}{\sqrt{mn}} \right). 
    \end{align*}
    The sub-Gaussian parameter of $\|z^T\mathbf{F}\|_1 + z^T g$ is 
    \begin{align*}
        \sqrt{\frac{(T - |\calK|)\xi^2}{mn} + \frac{|\calK|\xi^2}{m}} \lesssim \sqrt{\left(\frac{1-p}{mn} + \frac{p}{m} \right) \cdot T\xi^2 }.
    \end{align*}
    Therefore, Hoeffding's inequality implies that
    \[ f(z) \geq \Theta\left(\frac{(1-p)T \xi}{\sqrt{mn}} \right) \]
    holds with probability at least
    \[ 1 - \exp\left[ -\Theta\left( \frac{(1-p)^2 T}{1-p + np} \right) \right]. \]
    Choosing 
    \begin{align*}
        T \geq \Theta\Bigg[\max\Bigg\{ &\frac{1}{p}\log\left(\frac{1}{\delta}\right),\frac{1-p+np}{(1-p)^2}\log\left(\frac{1}{\delta}\right) \Bigg\}\Bigg]\\
            &\hspace{-5em}=\Theta\Bigg[\max\Bigg\{\frac{1}{p}\log\left(\frac{1}{\delta}\right),\frac{np}{(1-p)^2}\log\left(\frac{1}{\delta}\right) \Bigg\}\Bigg],
    \end{align*}    
    we have
    \[ \mathbb{P}\left[ f(z) \geq \Theta\left(\frac{(1-p)T \xi}{\sqrt{mn}} \right) \right] \geq 1 - \frac{\delta}2. \]
    Similarly, applying the discretization techniques, it is sufficient to choose $N$ points, where
    \begin{align*}
        &\quad \log(N)\\
        &= m \log\left[1 + \Theta\left( \frac{\sum_{i\notin\calK}\|u_i\|_2/\sqrt{n} + \sum_{i\in\calK}\|u_i\|_2 }{(1-p)T \xi / \sqrt{mn}} \right) \right]\\
        &\lesssim m \log\left[1 + \Theta\left( \frac{\sqrt{(1-p)T/n + pT}\xi \cdot \sqrt{\log(1/\delta)} }{(1-p)T \xi / \sqrt{mn}} \right) \right]\\
        &= m \log\left[1 + \Theta\left( \frac{\sqrt{1-p + np} \cdot \sqrt{\log(1/\delta)} }{(1-p)\sqrt{T/m}} \right) \right]\\
        &\leq m \log\left[ m \log\left(\frac1{\delta}\right)\right].
    \end{align*}
    Hence, the overall sample complexity is
    \[ T \geq \Theta\left[ nR_2 \left[m\log(nR_2) + \log\left(\frac{1}{\delta}\right)\right]\right], \]
    where we define
    \begin{align*}
        R_2 := \max\Bigg\{\frac{1}{np},\frac{p}{(1-p)^2}, \frac{m}{n} \Bigg\}.
    \end{align*}

    Combining the two steps, we get the conclusion of this theorem.

    \subsection{Proofs for Results in Appendix}

    \subsubsection{Proof of Lemma \ref{lem:support-1}}

        For the notational simplicity, we omit the $X$ in subscripts. Let
        \[ \eta := \frac{\tilde{\sigma}}{\sigma},\quad \delta := 1 - \frac{\tilde{\sigma}^4}{64\sigma^4}. \]
        Assume conversely that
        \[ \mathbb{P}(|X| \geq \eta\sigma) < 1 - \delta. \]
        Then, we can calculate that
        \begin{align*}
        \mathbb{E}(X^2) &= \int_0^\infty \theta^2 ~d\left[ -\mathbb{P}(|X| \geq \theta) \right]= \int_0^\infty 2\theta \mathbb{P}(|X| \geq \theta)~d\theta\\
        &\leq (\eta\sigma)^2 + \int_{\eta\sigma}^\infty 2\theta \min\left\{1 - \delta, 2\exp\left(-\frac{\theta^2}{2\sigma^2}\right)\right\}~d\theta\\
        &= (\eta\sigma)^2 + (1-\delta)\left[ (\sigma')^2 - (\eta\sigma)^2 \right]\\
        &\quad+ \int_{\sigma'}^\infty 2\theta \cdot2\exp\left(-\frac{\theta^2}{2\sigma^2}\right)~d\theta\\
        &= (\eta\sigma)^2 + (1-\delta)\left[ (\sigma')^2 - (\eta\sigma)^2 \right]\\
        &\quad+ 4\sigma^2\exp\left(-\frac{(\sigma')^2}{2\sigma^2}\right),
        \end{align*}
        where we define
        \[ \sigma' := \sqrt{2\sigma^2 \log\left( \frac{2}{1-\delta} \right)}. \]
        Rearranging the above inequality, we get
        \begin{align*}
            \eta^2 \cdot \delta + 2(1-\delta)\log\left( \frac{2}{1-\delta}\right) + 2(1-\delta) \geq \frac{\tilde{\sigma}^2}{\sigma^2}.
        \end{align*}
        Hence, it holds that
        \begin{align*} 
            \eta^2 &\geq \frac{1}{\delta}\left[ \frac{\tilde{\sigma}^2}{\sigma^2} - 2(1-\delta)\log\left( \frac{2}{1-\delta}\right) - 2(1-\delta) \right]\\
            &> 2\left[ \frac{\tilde{\sigma}^2}{\sigma^2} + 4(1-\delta)\log\left( \frac{1-\delta}2 \right) \right],
        \end{align*}
        where the second inequality holds because $\delta > 1/2$ and $\log[2 / (1-\delta)] > 1$. Using the fact that
        \[ (1 - \delta)\log\left(\frac{1-\delta}2\right) \geq -\sqrt{1-\delta} \geq -\frac{\tilde{\sigma}^2}{8\sigma^2}, \]
        we get
        \[ \eta^2 > \frac{\tilde{\sigma}^2}{\sigma^2}, \]
        which contradicts with the definition of $\eta$. Therefore, we have proved that
        \[ \mathbb{P}(|X| \geq \tilde{\sigma}) \geq \frac{\tilde{\sigma}^4}{64\sigma^4}. \]

\section*{References}
\bibliographystyle{IEEEtran}
\bibliography{main}

\begin{thebibliography}{10}
\providecommand{\url}[1]{#1}
\csname url@samestyle\endcsname
\providecommand{\newblock}{\relax}
\providecommand{\bibinfo}[2]{#2}
\providecommand{\BIBentrySTDinterwordspacing}{\spaceskip=0pt\relax}
\providecommand{\BIBentryALTinterwordstretchfactor}{4}
\providecommand{\BIBentryALTinterwordspacing}{\spaceskip=\fontdimen2\font plus
\BIBentryALTinterwordstretchfactor\fontdimen3\font minus \fontdimen4\font\relax}
\providecommand{\BIBforeignlanguage}[2]{{%
\expandafter\ifx\csname l@#1\endcsname\relax
\typeout{** WARNING: IEEEtran.bst: No hyphenation pattern has been}%
\typeout{** loaded for the language `#1'. Using the pattern for}%
\typeout{** the default language instead.}%
\else
\language=\csname l@#1\endcsname
\fi
#2}}
\providecommand{\BIBdecl}{\relax}
\BIBdecl

\bibitem{chen2012identification}
H.-F. Chen and L.~Guo, \emph{Identification and stochastic adaptive control}.\hskip 1em plus 0.5em minus 0.4em\relax Springer Science \& Business Media, 2012.

\bibitem{hazan2018spectral}
E.~Hazan, H.~Lee, K.~Singh, C.~Zhang, and Y.~Zhang, ``Spectral filtering for general linear dynamical systems,'' \emph{Advances in Neural Information Processing Systems}, vol.~31, 2018.

\bibitem{mania2019certainty}
H.~Mania, S.~Tu, and B.~Recht, ``Certainty equivalence is efficient for linear quadratic control,'' \emph{Advances in Neural Information Processing Systems}, vol.~32, 2019.

\bibitem{sarkar2020nonparametric}
T.~Sarkar, A.~Rakhlin, and M.~A. Dahleh, ``Nonparametric finite time {LTI} system identification,'' \emph{arXiv preprint arXiv:1902.01848}, 2019.

\bibitem{tsiamis2022statistical}
A.~Tsiamis, I.~Ziemann, N.~Matni, and G.~J. Pappas, ``Statistical learning theory for control: A finite sample perspective,'' \emph{arXiv preprint arXiv:2209.05423}, 2022.

\bibitem{Alan2022ControlBF}
\BIBentryALTinterwordspacing
A.~Alan, A.~J. Taylor, C.~R. He, A.~Ames, and G.~Orosz, ``Control barrier functions and input-to-state safety with application to automated vehicles,'' \emph{IEEE Transactions on Control Systems Technology}, vol.~31, pp. 2744--2759, 2022. [Online]. Available: \url{https://api.semanticscholar.org/CorpusID:249461776}
\BIBentrySTDinterwordspacing

\bibitem{Wang2017SafeLO}
\BIBentryALTinterwordspacing
L.~Wang, E.~A. Theodorou, and M.~Egerstedt, ``Safe learning of quadrotor dynamics using barrier certificates,'' \emph{2018 IEEE International Conference on Robotics and Automation (ICRA)}, pp. 2460--2465, 2017. [Online]. Available: \url{https://api.semanticscholar.org/CorpusID:35948052}
\BIBentrySTDinterwordspacing

\bibitem{KhansariZadeh2014LearningCL}
\BIBentryALTinterwordspacing
S.~M. Khansari-Zadeh and A.~Billard, ``Learning control lyapunov function to ensure stability of dynamical system-based robot reaching motions,'' \emph{Robotics Auton. Syst.}, vol.~62, pp. 752--765, 2014. [Online]. Available: \url{https://api.semanticscholar.org/CorpusID:14374268}
\BIBentrySTDinterwordspacing

\bibitem{simchowitz2018learning}
M.~Simchowitz, H.~Mania, S.~Tu, M.~I. Jordan, and B.~Recht, ``Learning without mixing: Towards a sharp analysis of linear system identification,'' in \emph{Conference On Learning Theory}.\hskip 1em plus 0.5em minus 0.4em\relax PMLR, 2018, pp. 439--473.

\bibitem{simchowitz2021naive}
M.~Simchowitz and D.~Foster, ``Naive exploration is optimal for online {LQR},'' in \emph{International Conference on Machine Learning}.\hskip 1em plus 0.5em minus 0.4em\relax PMLR, 2020, pp. 8937--8948.

\bibitem{zhang2021regret}
R.~Zhang, Y.~Li, and N.~Li, ``On the regret analysis of online {LQR} control with predictions,'' in \emph{2021 American Control Conference (ACC)}.\hskip 1em plus 0.5em minus 0.4em\relax IEEE, 2021, pp. 697--703.

\bibitem{ziemann2023tutorial}
I.~Ziemann, A.~Tsiamis, B.~Lee, Y.~Jedra, N.~Matni, and G.~J. Pappas, ``A tutorial on the non-asymptotic theory of system identification,'' 2023.

\bibitem{bakoAnalysisNonsmoothOptimization2016}
L.~Bako and H.~Ohlsson, ``Analysis of a nonsmooth optimization approach to robust estimation,'' \emph{Automatica}, vol.~66, pp. 132--145, Apr. 2016.

\bibitem{xuRobustnessRegularizationSupport2009}
H.~Xu, C.~Caramanis, and S.~Mannor, ``Robustness and {{Regularization}} of {{Support Vector Machines}}.'' \emph{Journal of machine learning research}, vol.~10, no.~7, 2009.

\bibitem{bakoClassOptimizationBasedRobust2017}
L.~Bako, ``On a {{Class}} of {{Optimization-Based Robust Estimators}},'' \emph{IEEE Transactions on Automatic Control}, vol.~62, no.~11, pp. 5990--5997, Nov. 2017.

\bibitem{bertsimasCharacterizationEquivalenceRobustification2018}
D.~Bertsimas and M.~S. Copenhaver, ``Characterization of the equivalence of robustification and regularization in linear and matrix regression,'' \emph{European Journal of Operational Research}, vol. 270, no.~3, pp. 931--942, Nov. 2018.

\bibitem{pesme2020online}
S.~Pesme and N.~Flammarion, ``Online robust regression via {SGD} on the {l-1} loss,'' \emph{Advances in Neural Information Processing Systems}, vol.~33, pp. 2540--2552, 2020.

\bibitem{dean2018sample}
S.~Dean, H.~Mania, N.~Matni, B.~Recht, and S.~Tu, ``On the sample complexity of the linear quadratic regulator,'' \emph{Foundations of Computational Mathematics}, vol.~20, no.~4, pp. 633--679, 2020.

\bibitem{mendelson2014learning}
S.~Mendelson, ``Learning without concentration,'' \emph{Journal of the ACM (JACM)}, vol.~62, no.~3, pp. 1--25, 2015.

\bibitem{li2021safe}
Y.~Li, S.~Das, J.~Shamma, and N.~Li, ``Safe adaptive learning-based control for constrained linear quadratic regulators with regret guarantees,'' \emph{arXiv preprint arXiv:2111.00411}, 2021.

\bibitem{fattahi2019learning}
S.~Fattahi, N.~Matni, and S.~Sojoudi, ``Learning sparse dynamical systems from a single sample trajectory,'' in \emph{2019 IEEE 58th Conference on Decision and Control (CDC)}.\hskip 1em plus 0.5em minus 0.4em\relax IEEE, 2019, pp. 2682--2689.

\bibitem{jedra2020finitetime}
Y.~Jedra and A.~Proutiere, ``Finite-time identification of stable linear systems optimality of the least-squares estimator,'' in \emph{2020 59th IEEE Conference on Decision and Control (CDC)}.\hskip 1em plus 0.5em minus 0.4em\relax IEEE, 2020, pp. 996--1001.

\bibitem{wagenmaker2020active}
A.~Wagenmaker and K.~Jamieson, ``Active learning for identification of linear dynamical systems,'' in \emph{Conference on Learning Theory}.\hskip 1em plus 0.5em minus 0.4em\relax PMLR, 2020, pp. 3487--3582.

\bibitem{feng2022learningnsp}
H.~Feng, B.~Yalcin, and J.~Lavaei, ``Learning of dynamical systems under adversarial attacks - null space property perspective,'' \emph{2023 American Control Conference (ACC)}, pp. 4179--4184, 2023.

\bibitem{wainwright_2019}
M.~J. Wainwright, \emph{High-Dimensional Statistics: A Non-Asymptotic Viewpoint}, ser. Cambridge Series in Statistical and Probabilistic Mathematics.\hskip 1em plus 0.5em minus 0.4em\relax Cambridge University Press, 2019.

\bibitem{han2021}
H.~Feng and J.~Lavaei, ``Learning of dynamical systems under adversarial attacks,'' in \emph{2021 60th IEEE Conference on Decision and Control (CDC)}, 2021, pp. 3010--3017.

\bibitem{chen2021bridging}
Y.~Chen, J.~Fan, C.~Ma, and Y.~Yan, ``Bridging convex and nonconvex optimization in robust pca: Noise, outliers, and missing data,'' \emph{Annals of statistics}, vol.~49, no.~5, p. 2948, 2021.

\bibitem{Farkas1902}
\BIBentryALTinterwordspacing
J.~Farkas, ``Theorie der einfachen ungleichungen.'' \emph{Journal für die reine und angewandte Mathematik}, vol. 124, pp. 1--27, 1902. [Online]. Available: \url{http://eudml.org/doc/149129}
\BIBentrySTDinterwordspacing

\bibitem{hovorka2002partitioning}
R.~Hovorka, F.~Shojaee-Moradie, P.~V. Carroll, L.~J. Chassin, I.~J. Gowrie, N.~C. Jackson, R.~S. Tudor, A.~M. Umpleby, and R.~H. Jones, ``Partitioning glucose distribution/transport, disposal, and endogenous production during ivgtt,'' \emph{American Journal of Physiology-Endocrinology and Metabolism}, vol. 282, no.~5, pp. E992--E1007, 2002.

\bibitem{compartment2018}
I.~Hajizadeh, M.~Rashid, and A.~Cinar, ``Integrating compartment models with recursive system identification,'' in \emph{2018 Annual American Control Conference (ACC)}, 2018, pp. 3583--3588.

\bibitem{hovorka2004nonlinear}
R.~Hovorka, V.~Canonico, L.~J. Chassin, U.~Haueter, M.~Massi-Benedetti, M.~O. Federici, T.~R. Pieber, H.~C. Schaller, L.~Schaupp, T.~Vering \emph{et~al.}, ``Nonlinear model predictive control of glucose concentration in subjects with type 1 diabetes,'' \emph{Physiological measurement}, vol.~25, no.~4, p. 905, 2004.

\end{thebibliography}

\begin{IEEEbiography}[{\includegraphics[width=1in,height=1.25in,clip,keepaspectratio]{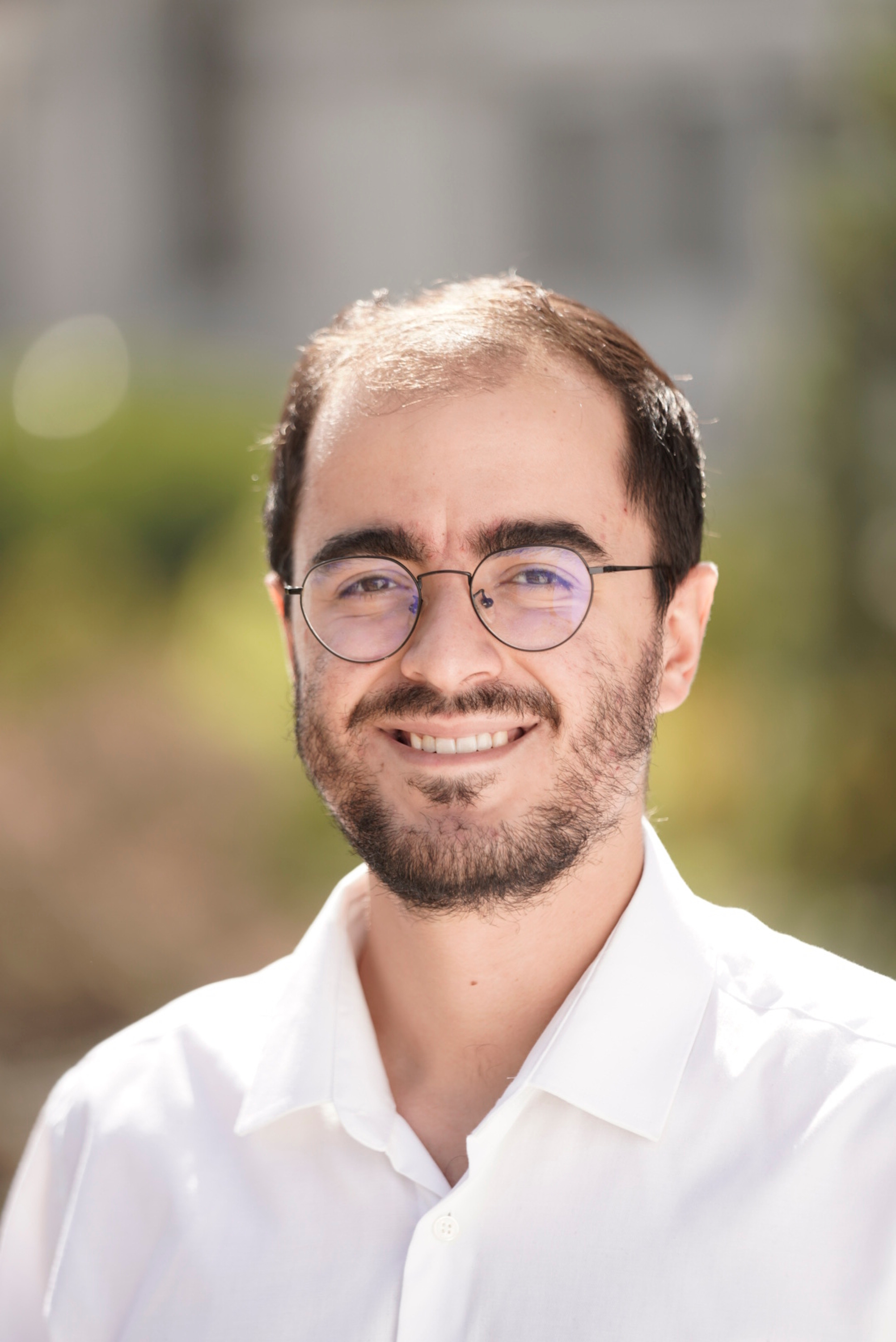}}]{Baturalp Yalcin,} is a Ph.D.candidate at the University of California, Berkeley in Industrial Engineering and Operations Research. His research interests include the landscape of optimization and adversarial learning problems. He holds a B.S. in Industrial Engineering from Bogazici University and an M.S. in Industrial Engineering and Operations Research from the University of California, Berkeley. 
\end{IEEEbiography}
\vspace{-1.5cm}
\begin{IEEEbiography}[{\includegraphics[width=1in,height=1.25in,clip,keepaspectratio]{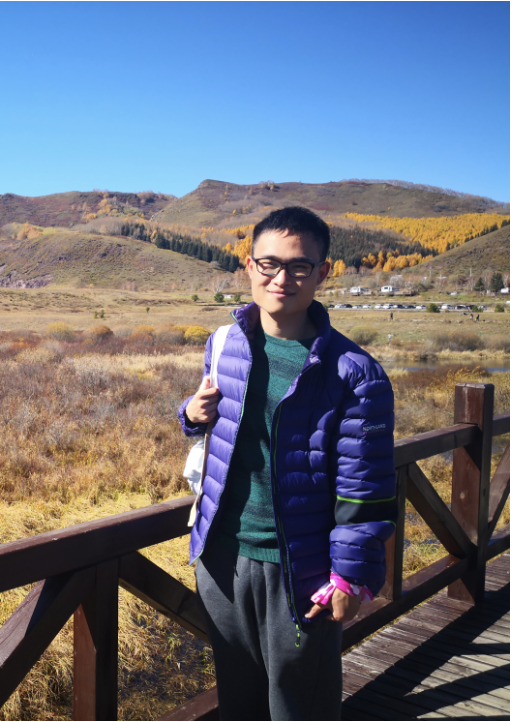}}]{Haixiang Zhang,} is a Ph.D.candidate at the University of California, Berkeley in Applied Mathematics. His research interests include nonconvex optimization, especially low-rank matrix optimization and optimization via simulation. He holds  B.S. in Computer Science and Technology and B.S. in Computational Mathematics from Peking University. He is the receipent of Two Sigma Ph.D. Fellowship.
\end{IEEEbiography}
\vspace{-1.5cm}
\begin{IEEEbiography}[{\includegraphics[width=1in,height=1.25in,clip,keepaspectratio]{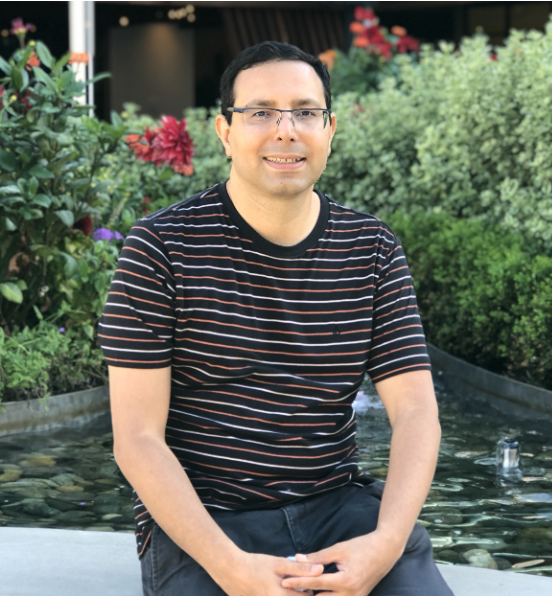}}]{Javad Lavaei,} received the Ph.D. degree in control and dynamical systems from the California Institute of Technology, Pasadena, CA, in 2011. He is an Associate Professor in the Department of Industrial Engineering and Operations Research, the University of California, Berkeley, Berkeley, CA. Dr. Lavaei received multiple awards, including the NSF CAREER Award, the Office of Naval Research Young Investigator Award, and the Donald P. Eckman Award.
\end{IEEEbiography}
\vspace{-1.5cm}
\begin{IEEEbiography}[{\includegraphics[width=1in,height=1.25in,clip,keepaspectratio]{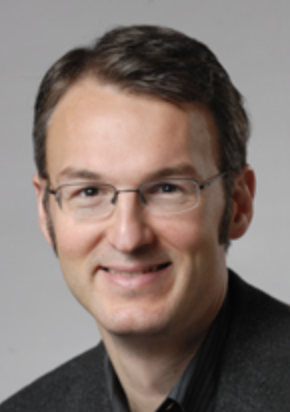}}]{Murat Arcak,} is currently a Professor with the Department of Electrical Engineering and 
Computer Sciences, University of California, Berkeley, CA, USA. His research interests include dynamical systems and control theory with applications to synthetic biology, multi-agent systems, and transportation. Prof. Arcak was a recipient of the NSF CAREER Award and the Donald P. Eckman Award.
\end{IEEEbiography}

\end{document}